\documentclass{article}

\usepackage{arxiv}

\usepackage{hyperref}
\usepackage{url}
\usepackage{graphicx}
\usepackage{wrapfig}
\usepackage{subcaption}
\usepackage{algorithm}%
\usepackage{algorithmicx}%
\usepackage{algpseudocode}%
\usepackage{varwidth}% http://ctan.org/pkg/varwidth
\usepackage{booktabs}
\usepackage{multirow}
\usepackage{comment}
\usepackage[numbers]{natbib}
\usepackage{amsmath}
\usepackage{amssymb}

\title{Adjoint sharding for very long context training of state space models}

\author{
    Xingzi Xu$\ ^{1,2}$\thanks {Work done during internship at Amazon.}
    \And
    Amir Tavanaei$\ ^{1}$
    \And 
    Kavosh Asadi$\ ^{1}$
    \And 
    Karim Bouyarmane$\ ^{1}$
    \AND\\
    $^{1}$\ Amazon \\ Seattle, WA 98109, USA \\ \texttt{\{xingzixu,atavanae,kavasadi,bouykari\}@amazon.com}
    \And \\
    $^{2}$ \ Duke University \\ Durham, NC 27708, USA
    \\\texttt{xingzi.xu@duke.edu}
}

\newtheorem{proposition}{{\bf Proposition}}

\newtheorem{proof}{Proof}

\newcommand{\A}{\mathcal{A}}
\newcommand{\B}{\mathcal{B}}
\newcommand{\C}{\mathcal{C}}

\newcommand{\R}{\mathbb{R}}

\newcommand{\T}{\mathbb{T}}

\newcommand{\ssm}{\mathrm{SSM}}

\newcommand{\vjp}{\mathrm{vjp}}
\newcommand{\diff}{\mathrm{d}}

\newcommand{\blambda}{\boldsymbol{\lambda}}

\floatevery{algorithm}{\setlength\hsize{10cm}}

\makeatletter
\newcommand\smaller{\@setfontsize\smaller\@viiipt\@viiiipt}
\makeatother

\begin{document}

\makeatletter
\renewcommand{\today}{December 31, 2024}
\makeatother
\date{\today}
\maketitle

\vspace{2em}  % Add extra space after the website section
\setcounter{footnote}{2}

\begin{abstract}
%Despite progress, efficiently training large language models (LLMs) in long contexts remains challenging. 
%Existing methods fall back to training LLMs with short contexts and use inference time techniques when evaluating on long contexts.
%Meanwhile, applications require training with long contexts, like enterprises fine-tuning pre-trained models on a local long context dataset or summarization tasks. 
%We propose adjoint sharding, a technique sharding gradient calculation during training to reduce memory requirements by orders of magnitude. Adjoint sharding is based on the adjoint method and computes equivalent gradients to backpropagation.
%We also propose truncated adjoint sharding to speed up the algorithm while maintaining performance.
%We provide a distributed version, and a paralleled version of adjoint sharding to further speed up training.
%Empirical results show the proposed adjoint sharding algorithm reduces memory usage. This removes the need for inference time techniques when evaluating on long sequences seen during training.
Despite very fast progress, efficiently training large language models (LLMs) in very long contexts remains challenging.
Existing methods fall back to training LLMs with short contexts (a maximum of a few thousands tokens in training) and use inference time techniques when evaluating on long contexts (above 1M tokens context window at inference). 
As opposed to long-context-inference, training on very long context input prompts is quickly limited by GPU memory availability and by the prohibitively long training times it requires on state-of-the-art hardware.
Meanwhile, many real-life applications require not only inference but also training/fine-tuning with long context on specific tasks. Such applications include, for example, augmenting the context with various sources of raw reference information for fact extraction, fact summarization, or fact reconciliation tasks. 
We propose adjoint sharding, a novel technique that comprises sharding gradient calculation during training to reduce memory requirements by orders of magnitude, making training on very long context computationally tractable. Adjoint sharding is based on the adjoint method and computes equivalent gradients to backpropagation.
We also propose truncated adjoint sharding to speed up the algorithm while maintaining performance.
We provide a distributed version, and a paralleled version of adjoint sharding to further speed up training.
Empirical results show the proposed adjoint sharding algorithm reduces memory usage by up to 3X with a 1.27B parameter large language model on 1M context length training. This allows to increase the maximum context length during training or fine-tuning of a 1.27B parameter model from 35K tokens to above 100K tokens on a training infrastructure composed of five AWS P4 instances. %It also removes the need for inference time techniques when evaluating on long sequences seen during training, improving accuracy by TK\% on example long-context task of fact extraction from multiple sources of context.
\footnote{Additional material for this paper can be found at: \url{https://adjoint-sharding.github.io}.}
\end{abstract}

\begin{figure}[hbt!]
%\vspace{-0pt}
  \begin{center}
    \includegraphics[width=0.9\textwidth, trim={0pt 10pt 5pt 10pt},clip]{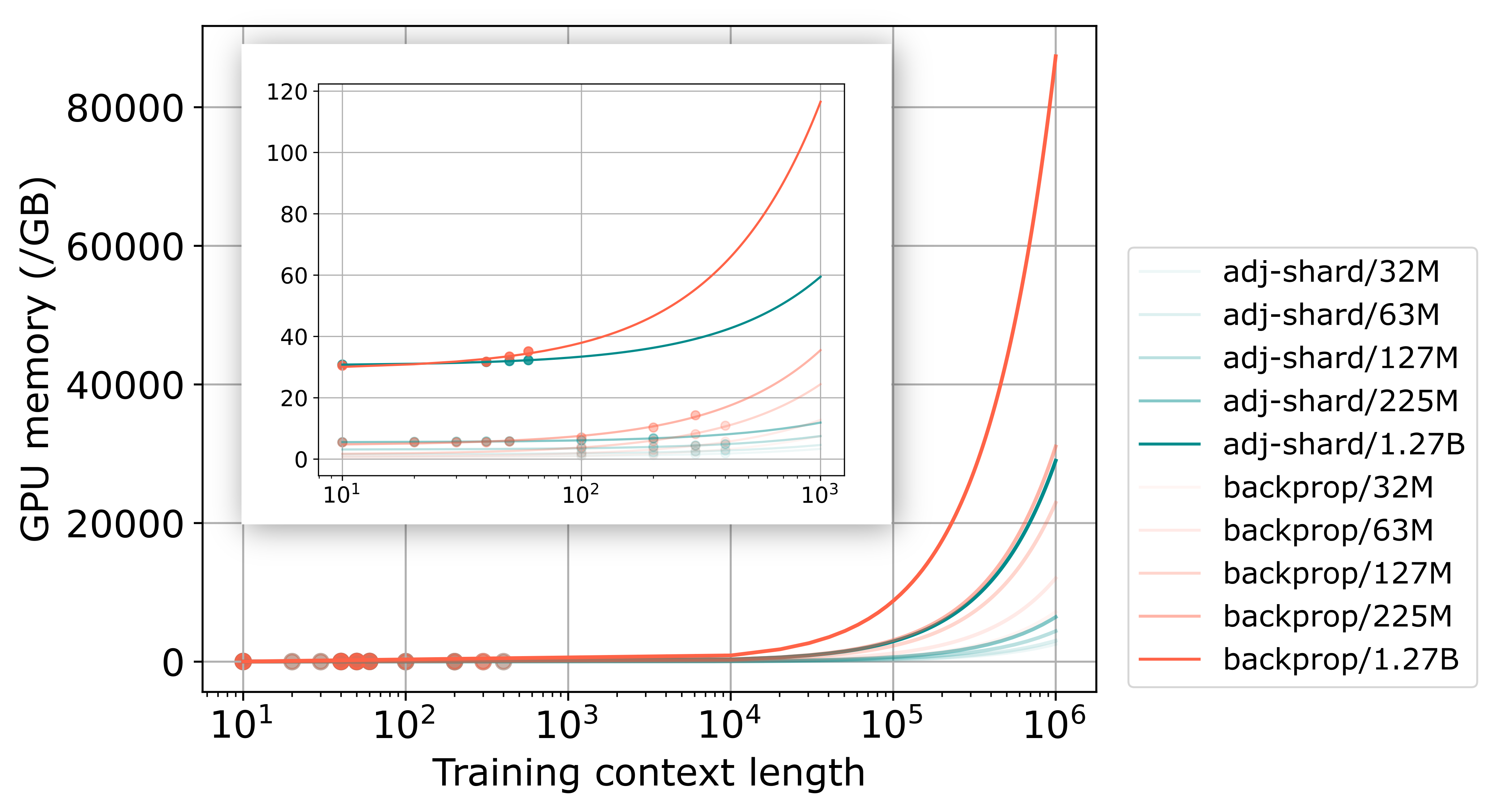}
    \caption{Compared to backpropagation (red lines), adjoint sharding (blue lines) significantly reduces memory requirements at training. Showing memory cost to train $32\mathrm{M}$, $63\mathrm{M}$, $127\mathrm{M}$, $225\mathrm{M}$, and $1.27\mathrm{B}$ parameter State Space Model (SSM) with batch size~$2$ and Adam optimizer on one GPU.}
    \label{fig:memoryReduction}
  \end{center}
  \vspace{-20pt}
\end{figure}

\section{Introduction}

\begin{comment}
\begin{wrapfigure}{r}{0.6\textwidth} 
\vspace{-70pt}
  \begin{center}
    \includegraphics[width=0.55\textwidth, trim={20pt 15pt 20pt 15pt},clip]{memoryReduction.pdf}
    \caption{Compared to backpropagation (red lines), adjoint sharding (blue lines) significantly reduces memory requirements. Showing the memory cost to train $32\mathrm{M}$, $63\mathrm{M}$, $127\mathrm{M}$, and $225\mathrm{M}$ parameter State Space Model (SSM) with batch size~$2$ and Adam optimizer on one GPU.}%\xx{run more exp}}
    \label{fig:memoryReduction}
  \end{center}
  \vspace{-20pt}
\end{wrapfigure}
\end{comment}

Foundation models are a new paradigm in artificial intelligence research focused on building large, general-purpose models that adapt to different tasks \cite{openai2024gpt4technicalreport2, dubey2024llama3herdmodels2, cai2024internlm2technicalreport2, pióro2024moemambaefficientselectivestate}. Extensive training on large datasets equips foundation models with broad capabilities, which are then fine-tuned on smaller datasets for specific applications. 
Foundation models commonly employ the transformer architecture \cite{vaswani2023attentionneed}. Despite the immense success, training transformer-based models requires memory growing quadratically with the context length $L$, limiting their applications on long context tasks \cite{li2024longcontextllmsstrugglelong}.
Researchers developed various techniques to conquer this problem, ranging from inference time context window expansion \citep{ding2024longropeextendingllmcontext, ding2024longrope}, IO-aware algorithms \citep{dao2022flashattentionfastmemoryefficientexact, dao2023flashattention2fasterattentionbetter, shah2024flashattention3fastaccurateattention}, and various linearly scaling language model architectures \citep{gu2024mambalineartimesequencemodeling, dao2024transformersssmsgeneralizedmodels, peng2023rwkvreinventingrnnstransformer, beltagy2020longformerlongdocumenttransformer}.
On another note, distributed learning enables training large models with a big number of GPUs, and efficient training methods like activation checkpointing, model/gradient sharding, and mixed-precision computing have further reduced the memory requirement of training a large model \cite{Verbraeken_2020, zhao2023pytorchfsdpexperiencesscaling, rajbhandari2020zeromemoryoptimizationstraining, micikevicius2018mixedprecisiontraining, herrmann2019optimalcheckpointingheterogeneouschains}.
However, current methodologies are entirely based on backpropagation and compute the gradient as a whole, inevitably requiring a memory growing rapidly with model size and context length \citep{damadi2023backpropagationalgorithmmathstudent}. Current sharding methods ignore the activations and only consider the model weights and optimizer states, constituting only a small fraction of the total memory cost \citep{sohoni2022lowmemoryneuralnetworktraining}. Activation checkpointing is among the limited techniques that consider activation values. Activation checkpointing offloads necessary intermediate states to the CPU and recompute them on the fly, trading compute time for memory reduction \citep{sohoni2022lowmemoryneuralnetworktraining, rajbhandari2020zero}. The substantial time required for offloading to the CPU hinders the effectiveness of activation checkpointing. 

We propose adjoint sharding to dissemble gradient computation of residual and/or recurrent based models to achieve orders of magnitude lower memory usage during training.

\begin{wrapfigure}{r}{0.5\textwidth} 
\vspace{-20pt}
  \begin{center}
    \includegraphics[width=0.45\textwidth, trim=25pt 10pt 15pt 41pt]{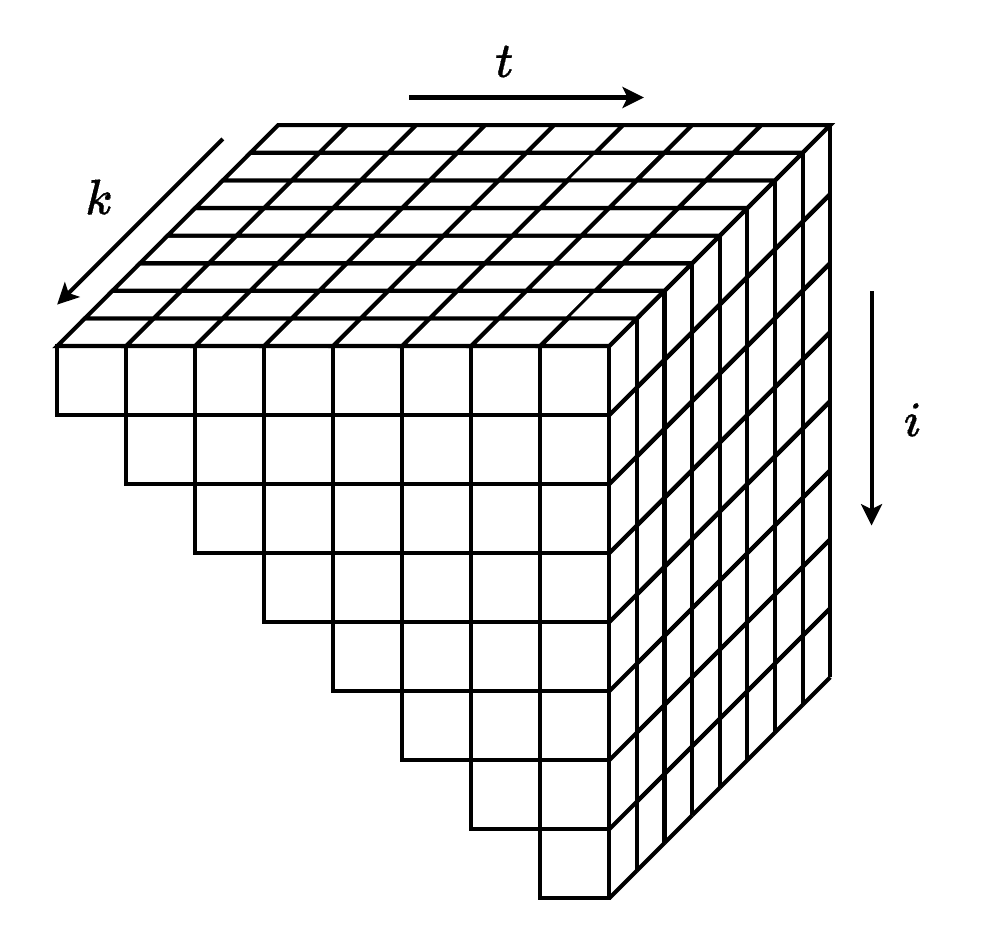}
    \caption{Adjoint sharding dissembles large models' gradient computations along the sequence dimension $t$ and the layer dimension $k$. When evaluating the gradient at time $t$, we perform $t$ vector-Jacobian products along the adjoint dimension $i$ for every layer indices $k$.}
    \label{fig:main}
  \end{center}
  \vspace{-50pt}
\end{wrapfigure}

\paragraph{Adjoint method} The adjoint sharding method is based on the adjoint method for recurrent models \citep{CAO2002171,adjointjohnson2007}. Given an optimization problem of a parametric recurrent forward process, the adjoint method is concerned with computations of the gradients regarding the process’s parameters. Backpropagation saves intermediate states to calculate gradients, whereas the adjoint method relies on a backward adjoint process to compute gradients. The adjoint method is a constant-memory optimization technique for dynamical systems \cite{chen2019neuralordinarydifferentialequations, xu2022characteristicneuralordinarydifferential}. In this paper, we are only concerned with the adjoint method for recurrent relations.
%The adjoint method has been popularized by the neural ordinary differential equation (neural ODE) community for its 

\paragraph{Vector-Jacobian product} Adjoint sharding dissembles the gradient computation of a large language model (LLM) into independent vector-Jacobian product (VJP) computations. By left-multiplying the Jacobian with a vector, it becomes unnecessary to compute the expensive Jacobian. Modern VJPs are as fast as a forward function call of the model, and can be thousands of times faster than Jacobian computations \cite{balestriero2021fastjacobianvectorproductdeep}. We speed up adjoint sharding by employing the VJPs.

\paragraph{Truncated adjoint sharding} Sharding the gradient computation allows us to prioritize the important gradients and disregard the rest, resulting in faster computation. We term this novel method truncated adjoint sharding, and empirically showcase its performance. 

\paragraph{Distributed and parallel computation} In addition, we have developed a distributed multi-GPU variant of adjoint sharding to further improve the scalability of LLM training. We also analyze the memory cost of parallel computation of adjoint sharding, opening up directions for massive speedups.

\paragraph{State-space models and residual networks} Residual networks (ResNets) are a commonly applied neural network structure. We illustrate adjoint sharding assuming a ResNet structure \citep{he2015deepresiduallearningimage}. State-space models (Mamba) have achieved performances on par with attention based models while possessing a linear scaling regarding the context length $L$, a polynomial speedup compared to the $L^2$ scaling of transformers \cite{vaswani2023attentionneed, gu2024mamba1}. 

\section{Related works}
\paragraph{Linear LLMs}
\citep{de2024griffinmixinggatedlinear, beck2024xlstmextendedlongshortterm, peng2023rwkvreinventingrnnstransformer} proposed LLM architectures with a linear inference time complexity. Each of them is formed by stacking $K$ residual layers together, where each layer has a recurrent relation. However, their temporal relationships are nonlinear, which limits the application of adjoint sharding to dissemble the gradients into independent vector-Jacobian products.

\paragraph{Backpropagation through time}
Applying the adjoint method for recurrent models leads to backpropagation through time (BPTT) \citep{bptt}. BPTT is a training algorithm developed for recurrent neural networks (RNNs). RNN models suffer from the exploding and vanishing gradient because of the $\prod_{j=i+1}^t\partial \mathbf{f}(\mathbf{x}^j, \mathbf{h}^{j-1},\mathbf{W}_{\mathbf{h}})/\partial \mathbf{h}^{j-1}$ term \citep{pascanu2013difficultytrainingrecurrentneural}. SSMs provide remedies with careful parameterization of the recurrent dynamics inspired by classical SSM theory \citep{fu2023hungryhungryhipposlanguage, gu2021combiningrecurrentconvolutionalcontinuoustime, gu2022trainhippostatespace, gupta2023simplifyingunderstandingstatespace, orvieto2023resurrectingrecurrentneuralnetworks, NEURIPS2020_c3581d21}. Linear temporal relations allow efficient evaluations of the model, while preserving universal approximation capabilities \citep{wang2023statespacemodelslayerwisenonlinearity}. By a similar token, truncated adjoint sharding can be seen as a more general version of the truncated backpropagation through time \citep{Jaeger2005ATO, tallec2017unbiasingtruncatedbackpropagationtime}. 

%To the best of our knowledge, we are the first to provide a principled view from the perspective of the adjoint method, to apply the adjoint method to SSMs, and to derive a distributed and paralleled training algorithm out of it.

\paragraph{Neural ordinary differential equations}
The adjoint method has also been applied to the optimization of continuous systems, especially the ordinary differential equations (ODEs) \citep{chen2019neuralordinarydifferentialequations, dupont2019augmentedneuralodes}. Optimizing neural ODEs with autograd requires backpropagating through numerical solvers along every step, using an unrealistic amount of memory. The adjoint method does not backpropagate through the operations of the solver and uses a constant amount of memory. However, applying the adjoint method for continuous systems requires solving a costly ODE initial value problem with dimensionality of the number of parameters.

\paragraph{Low memory training methods}
Researchers proposed various low memory training techniques to train big models in long contexts. ZERO provides data- and model-parallel training while retaining low communication volume, while eliminating memory redundancies \citep{rajbhandari2020zeromemoryoptimizationstraining}. PyTorch FSDP provides a streamline for model, gradient, and data parallelization \citep{zhao2023pytorchfsdpexperiencesscaling}. Activation checkpointing discards intermediate values during the forward step, and recompute on the fly during the training phase \citep{sohoni2022lowmemoryneuralnetworktraining}. CPU offloading scales large model training by offloading data and computations to the CPU, trading computing time for memory reduction \citep{ren2021zerooffloaddemocratizingbillionscalemodel}. Ring attention leverages the blockwise computation of self-attention and feedforward to distribute long sequences across multiple devices while fully overlapping the communication of key-value blocks with the computation of blockwise attention, enabling very-long context training of attention-based methods \citep{liu2023ringattentionblockwisetransformers, liu2024worldmodelmillionlengthvideo}. The proposed adjoint sharding distributes state-space model computations across multiple devices as well as multiple multi-GPU-instances (MIG) to enable very-long context training of state-space models.

\begin{figure}[hbt!]
%\vspace{-0pt}
  \begin{center}
    \includegraphics[width=0.8\textwidth, trim={0pt 5pt 5pt 5pt},clip]{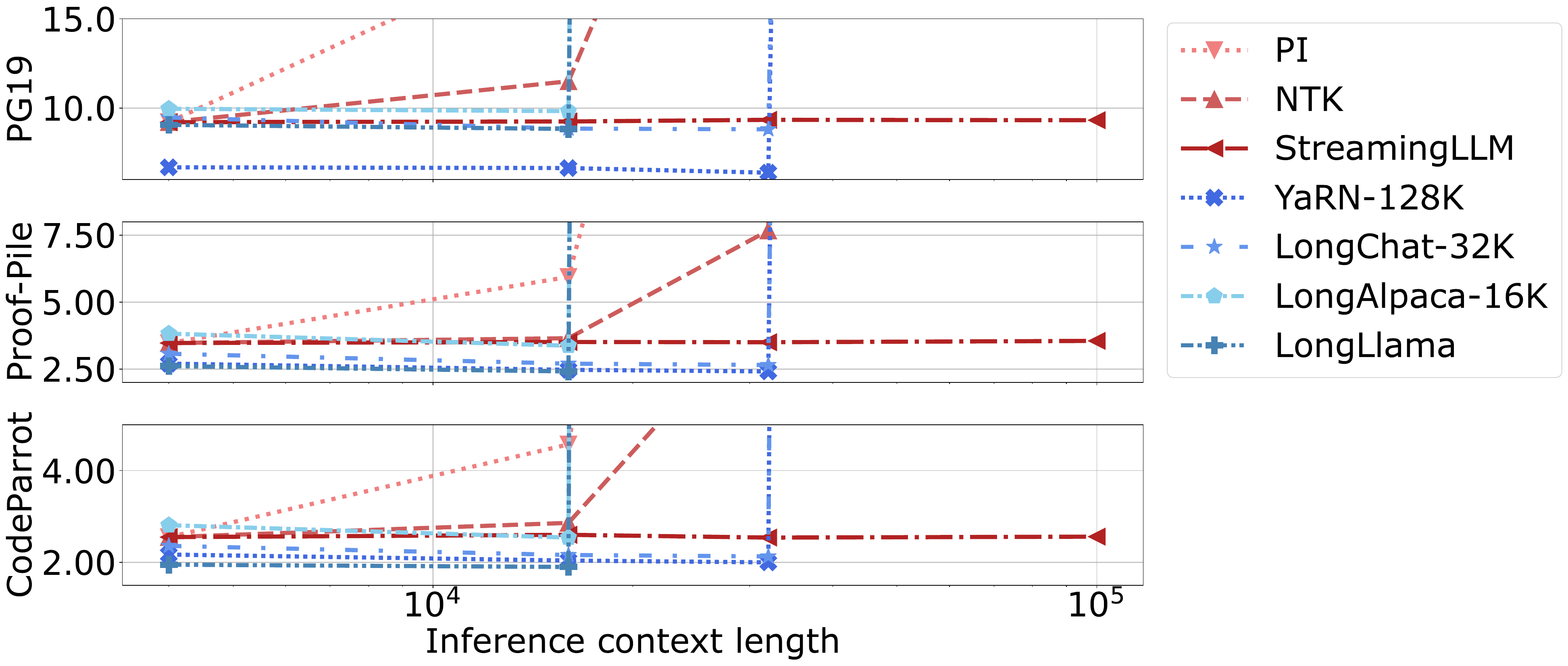}
    \caption{Lines in red are fine-tuning free methods and lines in blue are fine-tuning methods. Fine-tuning methods achieve better performances than fine-tuning free method but often suffer from out of memory issues \citep{chen2023extendingcontextwindowlarge, ntkReddit2023, xiao2024efficientstreaminglanguagemodels, longchat2023, chen2024longloraefficientfinetuninglongcontext, peng2023yarnefficientcontextwindow, zhang2024soaring4k400kextending, tworkowski2023focusedtransformercontrastivetraining}. Lower values are better across all three tasks.}
    \label{fig:pg19}
  \end{center}
  \vspace{-20pt}
\end{figure}

\paragraph{Context length extension methods}
Existing context length extension method separate into two classes. The first type is fine-tuning free methods, including Positional Interpolation (PI) \citep{chen2023extendingcontextwindowlarge}, the NTKAware Scale ROPE (NTK) \citep{ntkReddit2023}, and StreamingLLM \citep{xiao2024efficientstreaminglanguagemodels}. The second type is fine-tuning methods, including LongChat \citep{longchat2023}, LongAlpaca \citep{chen2024longloraefficientfinetuninglongcontext}, YaRN \citep{peng2023yarnefficientcontextwindow}, and LongLlama \citep{chen2024longloraefficientfinetuninglongcontext}. Additional methods such as activation beacon do tune a network seperate from the LLM \citep{zhang2024soaring4k400kextending}. As shown in \autoref{fig:pg19}, fine-tuning methods achieve better performances than that of fine-tuning free methods at lengths that they have been fine-tuned on. However, fine-tuning methods suffer from a high computational cost and require a potentially intractable amount of GPU memory during fine-tuning.

\section{Background}
We first give a concise introduction to the state-space models, the residual networks, and the adjoint method.
\subsection{State-space models}
\label{sec:ssm}
While our method generally applies to all recurrent models, we illustrate the idea using state-space models ($\ssm$s), which have shown performances at least on par with transformers at small to medium scale \citep{dao2024mamba2}. Given an input token sequence $\{\mathbf{x}_t\}_{t=1}^T$, the $\ssm$s first calculate the corresponding matrices $\mathbf{A}^t$, $\mathbf{B}^t$, and $\mathbf{C}^t$ to evolve the dynamics as follows:
\begin{align*}
    \mathbf{A}^t = \boldsymbol{\A}(\mathbf{x}^t);\; \mathbf{B}^t = \boldsymbol{\B}(\mathbf{x}^t);\; \mathbf{C}^t = \boldsymbol{\C}(\mathbf{x}^t). 
\end{align*}
The $\ssm$s evolve a latent dynamics $\mathbf{h}^t$, whose initial condition $\mathbf{h}^0$ is often assumed to be zero. With $\mathbf{h}^0$ and $\mathbf{A}^t,\,\mathbf{B}^t$ defined, the dynamics evolves as:
\begin{align*}
    \mathbf{h}^t = \mathbf{A}^t\mathbf{h}^{t-1}+\mathbf{B}^t\mathbf{x}^t.
\end{align*}
The matrices $\mathbf{C}^t$ then maps the latent dynamics $\mathbf{h}^t$ back to token space as $\mathbf{y}^t=\mathbf{C}^t \mathbf{h}^t$, with $\mathbf{y}^t$ being the predicted token at $t$. For a sequence of $T$ tokens, we denote:
\begin{align*}
    \mathbf{A} &= (\mathbf{A}^1,\mathbf{A}^2,\dots,\mathbf{A}^T),\;\mathbf{B} = (\mathbf{B}^1,\mathbf{B}^2,\dots,\mathbf{B}^T),\;\mathbf{C} = (\mathbf{C}^1,\mathbf{C}^2,\dots,\mathbf{C}^T),\\
    \mathbf{H} &= (\mathbf{h}^1,\mathbf{h}^2,\dots,\mathbf{h}^T),\;\mathbf{X} = (\mathbf{x}^1,\mathbf{x}^2,\dots,\mathbf{x}^T),\;\mathbf{Y} = (\mathbf{y}^1,\mathbf{y}^2,\dots,\mathbf{y}^T).
\end{align*}
In the most general case, we have $\mathbf{H}\in\R^{T\times N}, \mathbf{A}\in\R^{T\times N\times N}, \mathbf{B}\in\R^{T\times N\times P}, \mathbf{C}\in\R^{T\times P\times N}, \mathbf{X}\in\R^{T\times P}, \mathbf{Y}\in\R^{T\times P}$, where $N$ is the hidden state dimension, and $P$ is the input/output dimension. %\kb{introduce the $N$ and $P$ notations, (hidden state dimension and input/output dimension)}. 
We evolve the dynamics for $t=1,\dots, T$, and assume that $\mathbf{h}^0$ is a fixed and predefined constant. 

The input to an $\ssm$ is $\mathbf{X}$ and $\mathbf{h}^0$, and the output is $\mathbf{Y}$. We define $\ssm(\cdot)$ as performing the following five steps:
\begin{enumerate}
\item $\begin{aligned}[t]
    \{\mathbf{A}^t\}_{t=1}^T = \{\boldsymbol{\A}(\mathbf{x}^t)\}_{t=1}^T,
\end{aligned}$
\item $\begin{aligned}[t]
    \{\mathbf{B}^t\}_{t=1}^T = \{\boldsymbol{\B}(\mathbf{x}^t)\}_{t=1}^T,
\end{aligned}$
\item $\begin{aligned}[t]
    \{\mathbf{C}^t\}_{t=1}^T = \{\boldsymbol{\C}(\mathbf{x}^t)\}_{t=1}^T,
\end{aligned}$
\item $\begin{aligned}[t]
    \{\mathbf{h}^t\}_{t=1}^{T} = \{\mathbf{A}^t\mathbf{h}^{t-1}+\mathbf{B}^t\mathbf{x}^t\}_{t=1}^T;
\end{aligned}$
\item $\begin{aligned}[t]
    \{\mathbf{y}^t\}_{t=1}^{T} = \{\mathbf{C}^t\mathbf{h}^t\}_{t=1}^T.
\end{aligned}$
\end{enumerate}

The input to the five steps is $\mathbf{X}$, and the output is $\mathbf{Y}$. We can then write $\ssm(\mathbf{X})=\mathbf{Y}$.
%\ka{in 3.2 we are using the function SSM(y) but we have not defined it yet. Natural to define it here since you have all the ingredients that SSM(y) can be written in terms of.}
SSMs decrease the quadratic computational complexity with sequence length on transformers to linear and decrease the large inference-time memory requirements from the key-value cache. SSM-based models at a small to medium scale have shown performances on par with or better than transformer-based models. For instance, \citep{pióro2024moemambaefficientselectivestate, anthony2024blackmambamixtureexpertsstatespace} shows that SSM-based mixture-of-experts (MOE) model outperforms baseline transformer-based MOE model on model sizes as big as 2400M parameters. \citep{waleffe2024empiricalstudymambabasedlanguage} performed an extensive empirical study and found that while SSMs outperform transformers on various tasks, they underperform on tasks which require strong copying, in-context learning, or long-context reasoning abilities. \citep{waleffe2024empiricalstudymambabasedlanguage} also experimented with a SSM-transformer hybrid model, which outperforms transformers and is up to eight times faster when generating tokens at inference time. \citep{lieber2024jambahybridtransformermambalanguage} trained a 52B parameter model and further affirmed the hybrid models performances. 
%\xx{a paragraph on ssms' performances, and hippos etc}
\subsection{Residual Networks}
\label{sec:resnet}
In practice, we have $K$ $\ssm$s stacked together, and we have a large language head (LLH) $\Omega\in\R^{\T\times P}$, where $\T$ is the number of all possible tokens.  
To predict a token, we have $\mathbf{o}^t=\Omega \hat{\mathbf{y}}_K^t$. 
Define $(\mathbf{y}_K^1,\dots,\mathbf{y}_K^T)=\mathbf{Y}_K$, a ResNet computes $\mathbf{Y}_K$ as follows:
\begin{align*}
    (\mathbf{y}_K^1,\dots,\mathbf{y}_K^T) &= \mathbf{Y}_{K-1} + \ssm_K(\hat{\mathbf{Y}}_{K-1})\\ %\textrm{\ka{you see we have not defined this SSM() yet}}\\
    &= \mathbf{Y}_0 + \ssm_{1}(\hat{\mathbf{Y}}_0) + \dots + \ssm_K(\hat{\mathbf{Y}}_{K-1})\\
    &= \mathbf{Y}_0 + \sum_{k=1}^K \ssm_k(\hat{\mathbf{Y}}_{k-1})= \mathbf{Y}_0 + \sum_{k=1}^K \tilde{\mathbf{Y}}_{k},
\end{align*}
where $\hat{\mathbf{Y}}_k = (\hat{\mathbf{y}}_k^1,\dots,\hat{\mathbf{y}}_k^T)= (\mathrm{Norm}(\mathbf{y}_k^1),\dots,\mathrm{Norm}(\mathbf{y}_k^T))$ and 
$\ssm_k(\hat{\mathbf{Y}}_{k-1}) = \Tilde{\mathbf{Y}}_k$. Therefore, for a latent state at time $t$ we have $\mathbf{y}_K^t = \mathbf{y}_0^t+ \sum_{k=1}^K \tilde{\mathbf{y}}_k^t$. % \ka{isn't the normalization function taking the entire y as input? I guess depends on the definition?}. \ka{Also, this is the first time we use $\tilde Y$ so either use underbrace to define it in the last line, or define it prior to using it.} 
%\begin{align*}
%    \mathbf{y}_K^t = \mathbf{y}_0^t+\sum_{k=1}^K \ssm_k(\hat{\mathbf{y}}_{k-1}^t) = \mathbf{y}_0^t+ \sum_{k=1}^K \tilde{\mathbf{y}}_k^t.
%\end{align*}
%\ka{while the equality holds, writing it like that makes it look like $y^{t}_{K}$ is a function of tokens that come after $t$ since we need to compute the whole SSM and pick the $t$ index.}

ResNet has been the foundation of numerous modern networks, including the transformers, diffusion models, segmentation models, SSMs, and more \citep{He_2016_CVPR, guo2022attention, kirillov2023segment, peebles2023scalable}. ResNet’s residual structure allows for a separation between gradients of each layer by applying differentiation on summations.
%\xx{a paragraph on resnet's performances}
\subsection{Adjoint method}
\label{sec:adjoint}
The adjoint method is concerned with optimizing $\mathbf{y}(\mathbf{h}(\boldsymbol{\theta}),\boldsymbol{\theta})$ with respect to $\boldsymbol{\theta}$, where $\mathbf{h}(\boldsymbol{\theta})\in\R^P$ is the solution to $\mathbf{f}(\mathbf{h}(\boldsymbol{\theta}),\boldsymbol{\theta})=0$ \citep{CAO2002171}. To employ gradient based algorithms like the stochastic gradient descent (SGD) or the Adam, we compute the derivative of $\mathbf{y}$ regarding $\boldsymbol{\theta}\in\R^{|\boldsymbol{\theta}|}$:
%\ka{with respect to rather than regarding?}

\begin{equation}
    \frac{\diff \mathbf{y}}{\diff \boldsymbol{\theta}} = \frac{\partial \mathbf{y}}{\partial \boldsymbol{\theta}} + \frac{\partial \mathbf{y}}{\partial \mathbf{h}}\frac{\partial \mathbf{h}}{\partial \boldsymbol{\theta}},
\end{equation}
%To emphasis that the $\partial \mathbf{y}/\partial \boldsymbol{\theta}$ here does not depend on $\mathbf{h}$ and thereon, we denote $\partial \mathbf{y}/\partial \boldsymbol{\theta}$ as $\bpsi_{\boldsymbol{\theta}}$.

with $\diff$ being the total derivative, and $\partial$ being the partial derivative. The adjoint method converts computing $\diff \mathbf{y}/\diff \boldsymbol{\theta}$ to solving an adjoint equation. In our case, we need the adjoint method for recurrence relations, where $\mathbf{y}$ is given by $\mathbf{y}=\mathbf{y}^t\equiv \mathbf{y}(\mathbf{h}^t(\boldsymbol{\theta}),\boldsymbol{\theta})$, and $\mathbf{h}$ is given by
\begin{equation}
\label{eqn:adj}
\begin{cases}
    \mathbf{h}^0 &= \mathbf{b}(\boldsymbol{\theta}),\\
    \mathbf{h}^t &= \mathbf{f}(t,\mathbf{h}^{t-1},\boldsymbol{\theta}).
\end{cases}
\end{equation}
%\ka{this newly defined $f$ is annoying. not to mention that it takes $t$ as input which is weird. We can write $h$ in terms of what we already have defined, right?}
We have 
\begin{equation}
    \frac{\diff \mathbf{f}(t,\mathbf{h}^{t-1},\boldsymbol{\theta})}{\diff \boldsymbol{\theta}} = \frac{\partial \mathbf{f}(t,\mathbf{h}^{t-1},\boldsymbol{\theta})}{\partial \boldsymbol{\theta}} + \frac{\partial \mathbf{f}(t,\mathbf{h}^{t-1},\boldsymbol{\theta})}{\partial \mathbf{h}^{t-1}}\frac{\partial \mathbf{h}^{t-1}}{\partial \boldsymbol{\theta}}.
\end{equation}
%Again, to emphasis that the $\partial \mathbf{f}/\partial \boldsymbol{\theta}$ here does not depend on $\mathbf{h}$ and thereon, we denote $\partial \mathbf{f}/\partial \boldsymbol{\theta}$ as $\bphi_{\boldsymbol{\theta}}$.
%Denoting $\partial \mathbf{y}^t/\partial \mathbf{h}$, $\partial \mathbf{f}/\partial \mathbf{h}$ as $\mathbf{y}^t_{\mathbf{h}}$, $\mathbf{f}_{\mathbf{h}}$, the adjoint method states:
\begin{proposition}{\citep{CAO2002171}}
\label{prop:adjoint}
    When the states $\mathbf{h}$ are defined as \autoref{eqn:adj}, the gradient of $\mathbf{y}$ with respect to $\boldsymbol{\theta}$ is given as:
    \begin{equation}
    \begin{cases}
        \diff \mathbf{y}^t / \diff \boldsymbol{\theta} &= \partial \mathbf{y}^t/\partial \boldsymbol{\theta}+\blambda^0\mathbf{b}(\boldsymbol{\theta})+\sum_{i=1}^t\blambda^i\left(\partial \mathbf{f}(i,\mathbf{h}^{i-1},\boldsymbol{\theta})/\partial \boldsymbol{\theta}\right),\\%\textrm{\ka{you mean $b(\theta)?$}}
        \blambda^t &= \partial \mathbf{y}^t/\partial \mathbf{h}^t,\\
        \blambda^{i-1} &= \blambda^i\left(\partial \mathbf{f}(i,\mathbf{h}^{i-1},\boldsymbol{\theta})/ \partial \mathbf{h}^{i-1}\right).%\textrm{\ka{maybe better to use $i$ as index for this line?}}
    \end{cases}
    \end{equation}
    Equivalently, we have $\blambda^i=(\partial \mathbf{y}^t/\partial \mathbf{h}^t)\left(\prod_{j=t}^{i+1}\left(\partial \mathbf{f}(j,\mathbf{h}^{j-1},\boldsymbol{\theta})/ \partial \mathbf{h}^{j-1}\right)\right)$ \citep{adjointjohnson2007}.
\end{proposition}
%\ka{worth having a paragraph elaborating on why this equivalence is important. In particular when we compute the $\Lambda$s then we can compute the gradient in parallel. Maybe rewrite the paragraph below, start with the benefit of adjoint, and only then juxtapose with backpropagation.}

After computing adjoint states $\{\blambda^i\}_{i=0}^t$, the computation of the elements of $\blambda^i(\partial \mathbf{f}(i,\mathbf{h}^{i-1},\boldsymbol{\theta})/\partial\boldsymbol{\theta})$ are independent, allowing parallelism. This computation is a vector-Jacobian product ($\vjp$), with $\blambda^i$ as the vector and $\partial \mathbf{f}(i,\mathbf{h}^{i-1},\boldsymbol{\theta})/\partial\boldsymbol{\theta}$ as the Jacobian. $\vjp$s can be evaluated with the reverse-mode automatic differentiation and initializing the reverse phase with $\blambda^i$ \cite{baydin2018ad}. As each $\vjp$ only requires saving their corresponding computation graph, and can be disposed after the computation, we can compute $\vjp$s in parallel on modern GPUs. We will discuss this in more details in \autoref{sec:parallel}.
%If $\mathbf{f}$ is formed as a composition of $\boldsymbol{\Psi}$ functions $\{\mathbf{f}^{\psi}\}_{\psi=1}^{\boldsymbol{\Psi}}$ and $\mathbf{f}=\mathbf{f}^{\boldsymbol{\Psi}}\circ\mathbf{f}^{\boldsymbol{\Psi}-1}\circ\dots\circ\mathbf{f}^{1}$, and their corresponding inputs are $\boldsymbol{\Phi}=\{\bphi^{\psi}\}_{\psi=1}^{\boldsymbol{\Psi}}$, we can compute the $\vjp$ as
Adjoint sharding aims to use the adjoint method to replace backpropagation, which solves:
\begin{align*}
\label{eqn:backprop}
    \frac{\diff \mathbf{y}^t}{\diff \boldsymbol{\theta}} &= \frac{\partial \mathbf{y}^t}{\partial \boldsymbol{\theta}} + \frac{\partial \mathbf{y}^t}{\partial \mathbf{h}^t}\biggl( \frac{\partial \mathbf{f}(t,\mathbf{h}^{t-1},\boldsymbol{\theta})}{\partial \boldsymbol{\theta}}+\frac{\partial \mathbf{f}(t,\mathbf{h}^{t-1},\boldsymbol{\theta})}{\partial \boldsymbol{h}^{t-1}}\\
    &\biggl [\frac{\partial \mathbf{f}(t-1,\mathbf{h}^{t-2},\boldsymbol{\theta})}{\partial \boldsymbol{\theta}}+\frac{\partial \mathbf{f}(t-1,\mathbf{h}^{t-2},\boldsymbol{\theta})}{\partial \boldsymbol{h}^{t-2}} \biggl \{\frac{\partial \mathbf{f}(t-2,\mathbf{h}^{t-3},\boldsymbol{\theta})}{\partial \boldsymbol{\theta}}+\dots \biggl \} \biggl ] \biggl ).
\end{align*}
The backpropagation requires a sequential accumulation of the gradients, computing from the outmost layer inwards, therefore needs to save the computation graph for computations at all time $t$'s and creates memory bottlenecks. %\ka{are you sure that's how backprop is implemented for recurrent nets in pytorch? aside from that, everything up to here looks much better now}

\section{Adjoint sharding}
We now introduce the adjoint sharding technique. We first illustrate the method assuming only one layer of $\ssm$, and generalize to $K$ layers.
\subsection{Adjoint sharding for one SSM}
\label{sec:onessm}
Large scale neural networks are usually trained with the autograd framework \citep{baydin2018automaticdifferentiationmachinelearning, paszke2019pytorchimperativestylehighperformance}. However, this framework suffers from a high memory cost when used with networks of recurrent nature \citep{baydin2018automaticdifferentiationmachinelearning}. Although activation checkpointing has been developed, which discards part of the intermediate values and recomputes them later on the fly, the memory cost is still high \citep{herrmann2019optimalcheckpointingheterogeneouschains}.
We employ the adjoint method for recurrence relations to further reduce the memory cost, and more importantly, to break the temporal dependencies of activations and parallelize their computations.

Define $\theta = \langle \theta_{\boldsymbol{\A}}, \theta_{\boldsymbol{\B}}, \theta_{\boldsymbol{\C}}\rangle$ as $\boldsymbol{\A}$'s, $\boldsymbol{\B}$'s, and $\boldsymbol{\C}$'s parameters, for loss $l^t=l(\mathbf{y}^t)$, in the context of a single-layer $\ssm$, we prove:
%\ka{define $\theta = \langle A, B, C\rangle$}
\begin{proposition}
\label{prop:onessm}
    The gradient $\diff l^t/\diff \boldsymbol{\theta}$ is given as 
    \begin{equation}
    \frac{\diff l^t}{\diff\boldsymbol{\theta}} = \left[ \sum_{i=1}^{t}\vjp_{\boldsymbol{\A}^i}(\frac{\diff l^t}{\diff \mathbf{y}^t}\blambda^{t,i}\otimes \mathbf{h}^{i-1})\right ] \oplus \left [ \sum_{i=1}^{t}\vjp_{\boldsymbol{\B}^i}(\frac{\diff l^t}{\diff \mathbf{y}^t}\blambda^{t,i}\otimes \hat{\mathbf{x}}^i) \right ] \oplus \vjp_{\boldsymbol{\C}^t}(\frac{\diff l^t}{\diff \mathbf{y}^t}\otimes \mathbf{h}^t),
    \end{equation}
    where the adjoint state $\blambda^{t,\tau}=\mathbf{C}^t(\prod_{i=1}^{t-\tau}\mathbf{A}^{t+1-i})$, $\vjp_{\mathrm{Net}^i}(v)=v\cdot \mathrm{Net}_{\boldsymbol{\theta}}(\mathrm{Input}^i)$, with $\boldsymbol{\theta}$ being $\mathrm{Net}$'s parameters and $i$ being the index of $\mathrm{Input}$, $\otimes$ is the vector outer product, and $\oplus$ is vector concatenation.
\end{proposition}

The proof of proposition \ref{prop:onessm} is in section \ref{sec:proofOnessm}. The gradient for parameters of $\boldsymbol{\A}$, and $\boldsymbol{\B}$ are each separated into $\{\vjp_{\boldsymbol{\A}^i}(\frac{\diff l^t}{\diff \mathbf{y}^t}\blambda^{t,i}\otimes \mathbf{h}^{i-1})\}_{i=1}^t$, $\{\vjp_{\boldsymbol{\B}^i}(\frac{\diff l^t}{\diff \mathbf{y}^t}\blambda^{t,i}\otimes \hat{\mathbf{x}}^i\}_{i=1}^t$, and the gradient for parameters of $\boldsymbol{\C}$ only depend on inputs at time $t$. After computing the adjoint states, these $\vjp$ computations are separate from each other on both the network and the temporal level. 

\begin{figure}[hbt!]
%\vspace{-0pt}
  \begin{center}
    \includegraphics[width=0.8\textwidth, trim=0pt 15pt 15pt 10pt]{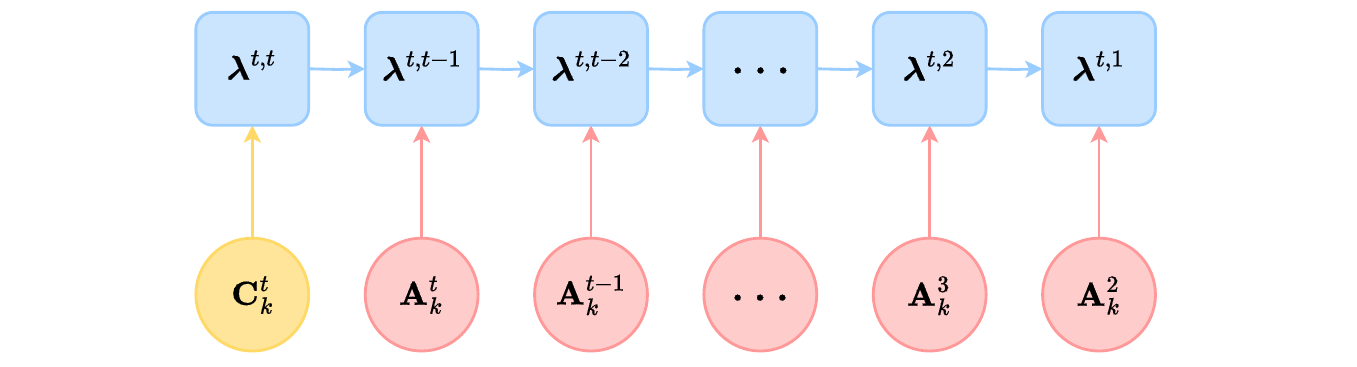}
    \caption{The adjoint states are computed sequentially backwards.}
    \label{fig:adjoint}
  \end{center}
  \vspace{-20pt}
\end{figure}

\subsection{Adjoint sharding for multiple SSMs}
We now generalize the results from \autoref{sec:onessm} to the general case of $K$ $\ssm$s concatenated together. As introduced in \autoref{sec:resnet}, the outputs of each $\ssm$ layer are added to the results of the last layer and normalized before it is fed into the next layer. Define the loss over all token predictions $L = \sum_{t=1}^T l^t$, using the residual structure we have
\begin{align*}
    \frac{\diff L}{\diff \boldsymbol{\theta}} = \sum_{t=1}^T\frac{\diff l^t}{\diff \mathbf{y}_K^t}\frac{\diff \mathbf{y}_K^t}{\diff \boldsymbol{\theta}}= \sum_{t=1}^T\frac{\diff l^t}{\diff \mathbf{y}_K^t}\frac{\diff (\mathbf{y}_0^t+\sum_{k=1}^K \tilde{\mathbf{y}}_k^t)}{\diff \boldsymbol{\theta}} = \sum_{t=1}^T\frac{\diff l^t}{\diff \mathbf{y}_K^t}\sum_{k=1}^K\frac{\diff \tilde{\mathbf{y}}_k^t}{\diff \boldsymbol{\theta}}.
\end{align*}
Combining with proposition \ref{prop:onessm}, we have
\begin{proposition}
\label{prop:multssm}
The gradient of the total loss $L$ with respect to the $\ssm$ parameters $\boldsymbol{\theta}$ is given as
\begin{equation}
\begin{split}
    \frac{\diff L}{\diff \boldsymbol{\theta}} &= \left(\sum_{t=1}^T\sum_{k=1}^K \sum_{i=1}^{t}\vjp_{\boldsymbol{\A}^i_k}(\frac{\diff l^t}{\diff \mathbf{y}_K^t}\blambda^{t,i}_k\otimes \mathbf{h}^{i-1}_k)\right) \\
    &\oplus \left(\sum_{t=1}^T\sum_{k=1}^K \sum_{i=1}^{t}\vjp_{\boldsymbol{\B}^i_k}(\frac{\diff l^t}{\diff \mathbf{y}_K^t}\blambda^{t,i}_k\otimes \hat{\mathbf{y}}^i_{k-1})\right) \\
    &\oplus \left(\sum_{t=1}^T\sum_{k=1}^K \vjp_{\boldsymbol{\C}^t_k}(\frac{\diff l^t}{\diff \mathbf{y}_K^t}\otimes \mathbf{h}^t_k)\right),
\end{split}
\end{equation}
where the input to $\vjp_{\boldsymbol{\C}^t_k}(\frac{\diff l^t}{\diff \mathbf{y}_K^t}\otimes \mathbf{h}^t_k),\, \vjp_{\boldsymbol{\A}^i_k}(\frac{\diff l^t}{\diff \mathbf{y}_K^t}\blambda^{t,i}_k\otimes \mathbf{h}^{i-1}_k), \,\mathrm{and}\,\vjp_{\boldsymbol{\B}^i_k}(\frac{\diff l^t}{\diff \mathbf{y}_K^t}\blambda^{t,i}_k\otimes \hat{\mathbf{y}}^i_{k-1})$ are computed with the k-th $\ssm$ and the $\hat{\mathbf{y}}_{k-1}^i=\mathrm{Norm}(\mathbf{y}_{k-2}^i+\ssm_{k-1}(\hat{\mathbf{Y}}_{k-2})^i)$ (the normalized output sequence of the (k-1)-th $\ssm$). The adjoint state at layer $k$ is defined as $\blambda^{t,\tau}_k=\mathbf{C}^t_k(\prod_{i=1}^{t-\tau}\mathbf{A}_k^{t+1-i})$.
\end{proposition}

\begin{wrapfigure}{r}{0.5\textwidth} %[hbt!]
\vspace{-10pt}
  \begin{center}
    \includegraphics[width=0.55\textwidth, trim=30pt 20pt 25pt 60pt]{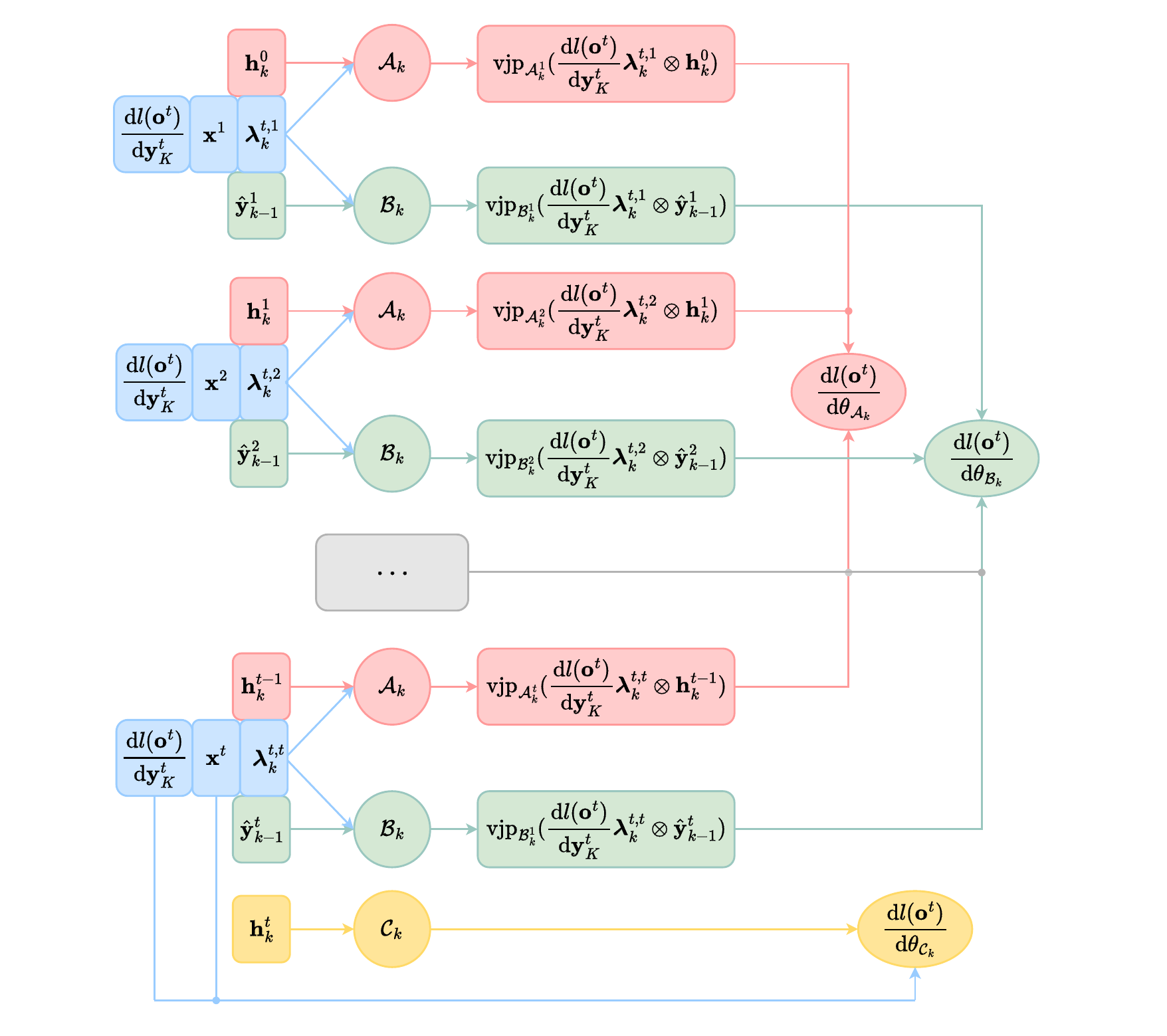}
    \caption{Computation schematic of $\diff l^t/\diff \boldsymbol{\theta}_{\boldsymbol{\A}_k}$, $\diff l^t/\diff \boldsymbol{\theta}_{\boldsymbol{\B}_k}$, and $\diff l^t/\diff \boldsymbol{\theta}_{\boldsymbol{\C}_k}$.}
    \label{fig:vjpABC}
  \end{center}
      \vspace{-50pt}
\end{wrapfigure}

We provide the proof to proposition \ref{prop:multssm} in section \ref{sec:proofMultssm}. Define $\boldsymbol{\Lambda}_k^t=\{\blambda^{t,\tau}_k\}_{\tau=1}^{t}$, proposition \ref{prop:multssm} shows that the gradients of each network’s parameters computed with each token only correlate through the adjoint states $\{\boldsymbol{\Lambda}_k^t\}_{k,t=1,1}^{K,T}$. The adjoint states can be easily computed after a forward pass. The adjoint states can also be computed on the fly in the gradient computation phase, as it only depends on $\mathbf{C}_k^t$ and $\mathbf{A}_k^t$ and has no dependencies on the network Jacobians regarding the network parameters. The adjoint sharding method breaks down the backpropagation computation both layer-wise and token-wise into foundational $\vjp$ computations that do not have any dependencies on each other. 

We show a schematic of the computations to $\diff l^t/\diff \boldsymbol{\theta}_{\boldsymbol{\A}_k}$, $\diff l^t/\diff \boldsymbol{\theta}_{\boldsymbol{\B}_k}$, and $\diff l^t/\diff \boldsymbol{\theta}_{\boldsymbol{\C}_k}$ in \autoref{fig:vjpABC} and a schematic for computing the adjoint states in \autoref{fig:adjoint}.

\subsection{Truncated adjoint sharding}

One limitation of adjoint sharding is that the number of $\vjp$s performed increases polynomially regarding the number of tokens $T$. In particular, adjoint sharding computes the $\vjp$ for $\boldsymbol{\A_k}$ and $\boldsymbol{\B_k}$ $(1+T)T/2$ times, and for $\boldsymbol{\C_k}$ $T$ times. When training large networks with many layers and long context length $T$, applying adjoint sharding becomes computationally expensive. 
We propose truncated adjoint sharding, with which we argue that we can get similar results by computing a linearly growing number of $\vjp$s, and empirically showcase its performance. 

Attention mechanisms have suffered from the $\mathcal{O}(T^2)$ complexities arising from the self-attention structure \citep{vaswani2023attentionneed}. To enable training with longer context lengths, global-local attention has been proposed, where we divide the contexts into sections, and compute the attention between sections rather than tokens \citep{yang2021focal}. \citep{tallec2017unbiasingtruncatedbackpropagationtime} proposed truncated backpropagation through time (T-BPTT) to avoid gradient explosion/vanishing when training with long contexts by only counting a fixed number of state transitions. Here, inspired by global-local attention and T-BPTT, instead of computing the full gradient given in \autoref{eqn:resnet}, we propose to train the $\ssm$s to depend on up to $\Bar{T}$ states:
\begin{equation}
\label{eqn:capped}
\begin{split}
    \frac{\diff L}{\diff \boldsymbol{\theta}} &= \left(\sum_{t=1}^T\sum_{k=1}^K \vjp_{\boldsymbol{\C}^t_k}(\frac{\diff l^t}{\diff \mathbf{y}_K^t}\otimes \mathbf{h}^t_k) \right) \\
     &\oplus \left(\sum_{t=1}^{\Bar{T}}\sum_{k=1}^K \sum_{i=1}^{t}\vjp_{\boldsymbol{\A}^i_k}(\frac{\diff l^t}{\diff \mathbf{y}_K^t}\blambda^{t,i}_k\otimes \mathbf{h}^{i-1}_k) + \sum_{t=\Bar{T}+1}^{T}\sum_{k=1}^K \sum_{i=t+1-\Bar{T}}^{t}\vjp_{\boldsymbol{\A}^i_k}(\frac{\diff l^t}{\diff \mathbf{y}_K ^t}\blambda^{t,i}_k\otimes \mathbf{h}^{i-1}_k)\right) \\
     &\oplus \left(\sum_{t=1}^{\Bar{T}}\sum_{k=1}^K \sum_{i=1}^{t}\vjp_{\boldsymbol{\B}^i_k}(\frac{\diff l^t}{\diff \mathbf{y}_K^t}\blambda^{t,i}_k\otimes \hat{\mathbf{y}}^i_{k-1}) + \sum_{t=\Bar{T}+1}^{T}\sum_{k=1}^K \sum_{i=t+1-\Bar{T}}^{t}\vjp_{\boldsymbol{\B}^i_k}(\frac{\diff l^t}{\diff \mathbf{y}_K^t}\blambda^{t,i}_k\otimes \hat{\mathbf{y}}^i_{k-1}\right)
\end{split}
\end{equation}

\begin{wrapfigure}{r}{0.5\textwidth} 
\vspace{-20pt}
  \begin{center}
    \includegraphics[width=0.5\textwidth, trim={0pt 10pt 0pt 10pt}, clip]{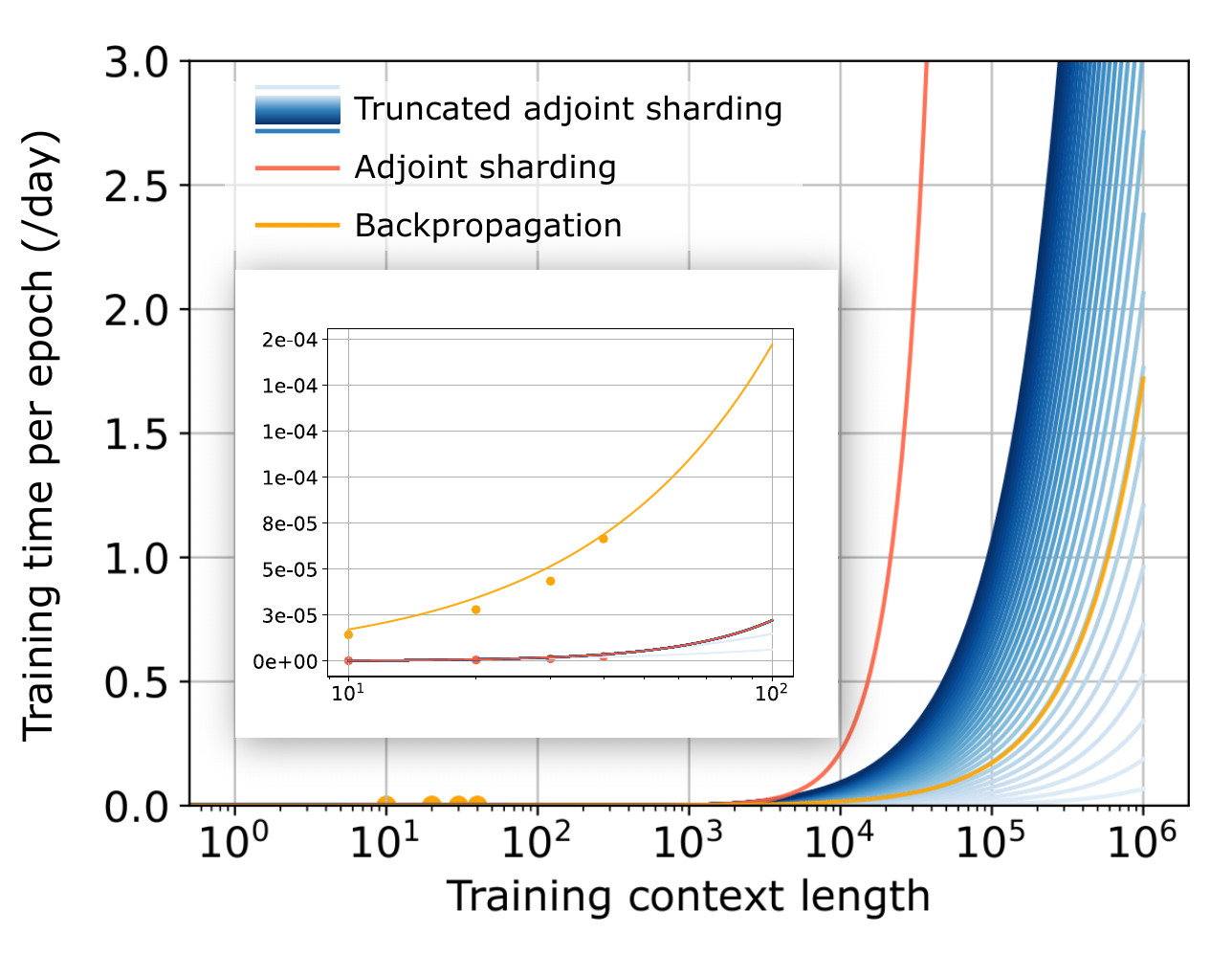}
    \caption{Training time (/day) per epoch comparison for adjoint sharding, truncated adjoint sharding, and backpropagation with different context lengths. Assumed a 100-layer $\ssm$-ResNet model, a 280x acceleration for adjoint sharding from parallel computing (achievable with five Amazon P4 instances), and $\Bar{T}$ from $15$ to $2500$.}
    \label{fig:tCapped}
  \end{center}
  \vspace{-20pt}
\end{wrapfigure}

As shown in \autoref{eqn:capped} above, we perform the same computations for $t=1,\dots,\Bar{T}$ as before, and only perform the $\vjp$s back to the last $\Bar{T}$ states for $t>\Bar{T}$. With truncated adjoint sharding, we perform $\Bar{T} T+ \Bar{T}(\Bar{T}-1)/2$ $\vjp$s, which grows linearly. We show the number of $\vjp$s performed with and without truncated adjoint sharding in \autoref{fig:tCapped}. When $\Bar{T}=2000$, truncated adjoint sharding reduces  $64\%$ of the $\vjp$s when training with a context length of $10\mathrm{K}$.

\begin{comment}
\begin{wrapfigure}{r}{0.5\textwidth} 
\vspace{-20pt}
  \begin{center}
    \includegraphics[width=0.5\textwidth, trim={0pt 0pt 0pt 40pt}, clip]{tCapped.pdf}
    \caption{Number of vector-Jacobian products performed with and without truncated adjoint sharding. Plotted with $\Bar{T}$ from $100$ to $2000$.}
    \label{fig:tCapped}
  \end{center}
  \vspace{-40pt}
\end{wrapfigure}
\end{comment}

The essence of the truncated adjoint sharding method is that we only explicitly count gradients related to the last $\Bar{T}$ states. As each state depends on its prior state, states still implicitly depend on all their prior states. We leave investigation of $\Bar{T}$'s impact on performances for future works.

%\xx{We perform an ablation study on $\Bar{T}$ to analyze its impact on the training performances. }

\subsection{Distributed training}
We now discuss how to distribute the storage and compute of the adjoint sharding method, assuming that we have $\Upsilon$ GPUs. Given the networks $\{\A_k, \B_k, \C_k\}_{k=1}^K$, initial tokens $\{\hat{\mathbf{y}}_0^t\}_{t=1}^T=\{\mathrm{Norm}(\mathbf{x}^t)\}_{t=1}^T$, and initial conditions $\{\mathbf{h}^0_k\}_{k=1}^K$ (usually set to $\mathbf{0}$), we can call algorithm \ref{alg:forward} to get all necessary vectors for computing the gradient with adjoint sharding. 

\begin{algorithm}
\caption{Forward step in evaluation mode on a distributed system}\label{alg:forward}
\begin{algorithmic}[1]
    \State \text{\bf Inputs: $\{\hat{\mathbf{y}}_0^t\}_{t=1}^T$, $\{\mathbf{h}^0_k\}_{k=1}^K$, $\{\A_k, \B_k, \C_k\}_{k=1}^K$, $\Omega$}
    \State \text{On devices $\upsilon=1,\dots,\Upsilon$, in parallel \textbf{do}}
        \For{SSM model index $k=(\upsilon-1)(K//\Upsilon)+1,\dots,\upsilon(K//\Upsilon)$}
            \For{Time step index $t=1,\dots,T$}
                \State Compute: $\mathbf{A}_k^t = \A_k(\hat{\mathbf{y}}_{k-1}^t)$;  $\mathbf{B}_k^t = \B_k(\hat{\mathbf{y}}_{k-1}^t)$; $\mathbf{C}_k^t = \C_k(\hat{\mathbf{y}}_{k-1}^t)$; $\mathbf{h}_k^t = \mathbf{A}_k^t\mathbf{h}_k^{t-1}+\mathbf{B}_k^t\hat{\mathbf{y}}_{k-1}^t$; $\mathbf{y}_k^t = \mathbf{C}_k^t\mathbf{h}_k^t$.
                \State Compute: $\mathbf{y}_k^t = \mathbf{y}_{k-1}^t+\tilde{\mathbf{y}}_k^t$.
                \State Compute: $\hat{\mathbf{y}}_k^t = \mathrm{Norm}(\mathbf{y}_k^t)$.
                %\State Store: $A_k^t$, $C_k^t$, $h_k^t$, $y_k^t$.
            \EndFor
        \EndFor
        \State Store: $\{\mathbf{h}_k^t\}_{(t,k)=(1,(\upsilon-1)(K//\Upsilon)+1)}^{T, \upsilon(K//\Upsilon)}$, $\{\mathbf{C}_k^t\}_{(t,k)=(1,(\upsilon-1)(K//\Upsilon)+1)}^{T, \upsilon(K//\Upsilon)}$, $\{\hat{\mathbf{y}}_k^t\}_{(t,k)=(1,(\upsilon-1)(K//\Upsilon))}^{T, \upsilon(K//\Upsilon)-1}$, $\{\mathbf{A}_k^t\}_{(t,k)=(2,(\upsilon-1)(K//\Upsilon)+1)}^{T, \upsilon(K//\Upsilon)}$ on device $\upsilon$.
        \State Pass: $\{\mathbf{y}_{\upsilon(K//\Upsilon)-1}^t\}_{t=1}^T$, $\{\hat{\mathbf{y}}_{\upsilon(K//\Upsilon)-1}^t\}_{t=1}^T$ to device $\upsilon+1$
    %\EndFor
    \For{Time step index $t=1,\dots,T$} 
        \State Compute: $\{\mathbf{o}^t=\Omega \mathbf{y}_K^t\}_{t=1}^T$, $\{l(\mathbf{o}^t)\}$, $\{\frac{\diff l(\mathbf{o}^t)}{\diff \mathbf{y}_K^t}\}_{t=1}^T$.
        %\State Store: $\{\frac{\diff l(\mathbf{o}^t)}{\diff y_K^t}\}_{t=1}^T$.
    \EndFor
    \State Store: $\{\frac{\diff l(\mathbf{o}^t)}{\diff \mathbf{y}_K^t}\}_{t=1}^T$ on all $\Upsilon$ devices.
\end{algorithmic}
\end{algorithm}

\begin{algorithm}
\caption{Evaluating adjoint states for token index $t$ and ResNet index $k$ with truncated adjoint sharding $\Bar{T}$}\label{alg:capped_adjoint}
\begin{algorithmic}[1]
    \State \text{\bf Inputs: $t$, $k$, $\Bar{T}$, $\mathbf{C}_k^t$, $\{\mathbf{A}^i_k\}_{i=t+2-\Bar{T}}^t$}
    \State \text{Initialize adjoint state }$\blambda_k^{t,t}=\mathbf{C}_k^t$
    \State \text{Compute: intermediate values:} \\$\boldsymbol{\zeta}^{\Bar{T}}=(\mathbf{A}_k^t\mathbf{A}_k^{t-1}\dots \mathbf{A}_k^{t+2-\Bar{T}}, \mathbf{A}_k^t\mathbf{A}_k^{t-1}\dots \mathbf{A}_k^{t+3-\Bar{T}},\dots,\mathbf{A}_k^t\mathbf{A}_k^{t-1},\mathbf{A}_k^t,\mathbb{I})$.
    \State \text{Compute: adjoint states} $\Bar{\boldsymbol{\Lambda}}_k^{\Bar{T}}=(\blambda_k^{t,t+1-\Bar{T}},\blambda_k^{t,t+2-\Bar{T}},\dots,\blambda_k^{t,t})=\mathbf{C}_k^t\boldsymbol{\zeta}^{\Bar{T}}$.
    %\State \text{Release} $\zeta$.
    \State \text{\bf Return: $\Bar{\boldsymbol{\Lambda}}_k^{\Bar{T}}$}.
\end{algorithmic}
\end{algorithm}

\begin{algorithm}
\caption{Evaluating the $\vjp$'s for token index $t$ and ResNet index $k$ with truncated adjoint sharding $\Bar{T}$}\label{alg:capped_vjp}
\begin{algorithmic}[1]
    \State \text{\bf Inputs: $t$, $k$, $\Bar{T}$, $\frac{\diff l(\mathbf{o}^t)}{\diff \mathbf{y}_K^t}$, $\{\mathbf{h}_k^i\}_{i=t-\Bar{T}}^{t}$, $\mathbf{C}_k^t$, $\{\mathbf{y}^i_{k-1}\}_{i=t+1-\Bar{T}}^{t}$, $\{\mathbf{A}^i_k\}_{i=t+2-\Bar{T}}^t$}
    \State \text{Call alg. \ref{alg:capped_adjoint} to compute $\{\blambda_k^{t,i}\}_{i=t+1-\Bar{T}}^t$}
    \State \text{Compute: $\frac{\diff l(\mathbf{o}^t)}{\diff \mathbf{y}_K^t}\otimes \mathbf{h}^t_k$, $\{\frac{\diff l(\mathbf{o}^t)}{\diff \mathbf{y}_K^t}\blambda^{t,i}_k\otimes \mathbf{h}^{i-1}_k\}_{i=t+1-\Bar{T}}^{t}$, $\{\frac{\diff l(\mathbf{o}^t)}{\diff \mathbf{y}_K^t}\blambda^{t,i}_k\otimes \hat{\mathbf{y}}^i_{k-1}\}_{i=t+1-\Bar{T}}^{t}$}
    \State \text{Compute:{\scriptsize $\left (\vjp_{\mathbf{C}^t_k}(\frac{\diff l(\mathbf{o}^t)}{\diff \mathbf{y}_K^t}\otimes \mathbf{h}^t_k),\, \sum_{i=t+1-\Bar{T}}^{t}\vjp_{\mathbf{A}^i_k}(\frac{\diff l(\mathbf{o}^t)}{\diff \mathbf{y}_K^t}\blambda^{t,i}_k\otimes \mathbf{h}^{i-1}_k),\,\sum_{i=t+1-\Bar{T}}^{t}\vjp_{\mathbf{B}^i_k}(\frac{\diff l(\mathbf{o}^t)}{\diff \mathbf{y}_K^t}\blambda^{t,i}_k\otimes \hat{\mathbf{y}}^i_{k-1}) \right )$}}
    \State \text{\bf Return: {\scriptsize $\left( \vjp_{\mathbf{C}^t_k}(\frac{\diff l(\mathbf{o}^t)}{\diff \mathbf{y}_K^t}\otimes \mathbf{h}^t_k),\, \sum_{i=t+1-\Bar{T}}^{t}\vjp_{\mathbf{A}^i_k}(\frac{\diff l(\mathbf{o}^t)}{\diff \mathbf{y}_K^t}\blambda^{t,i}_k\otimes \mathbf{h}^{i-1}_k),\,\sum_{i=t+1-\Bar{T}}^{t}\vjp_{\mathbf{B}^i_k}(\frac{\diff l(\mathbf{o}^t)}{\diff \mathbf{y}_K^t}\blambda^{t,i}_k\otimes \hat{\mathbf{y}}^i_{k-1}) \right)$}}
\end{algorithmic}
\end{algorithm}

\begin{algorithm}
\caption{Evaluating $\frac{\diff L}{\diff \boldsymbol{\theta}}$ with truncated adjoint sharding $\Bar{T}$ on $\Upsilon$ devices}\label{alg:dist_capped_grad}
\begin{algorithmic}[1]
    \State \text{\bf Inputs: $\{\mathbf{y}_0^t\}_{t=1}^T$, $\{\mathbf{h}^0_k\}_{k=1}^K$, $\{\A_k, \B_k, \C_k\}_{k=1}^K$, $\Omega$, $\Bar{T}$, $\Upsilon$}
    \State \text{Call alg. \ref{alg:forward} for } $\{\mathbf{A}_k^t, \mathbf{C}_k^t, \mathbf{h}_k^t, \hat{\mathbf{y}}_k^t\}_{(t,k)=(1,1)}^{(T,K)}, \{\frac{\diff l(\mathbf{o}^t)}{\diff \mathbf{y}_K^t}\}_{t=1}^T$ and saved on each GPU device.
    \State On each device $\upsilon$, in parallel \textbf{do}
    \State \text{Initialize gradient } $\frac{\diff L}{\diff \boldsymbol{\theta}}$
    \For {Time step index $t=1,\dots,\Bar{T}$, layer index $k=(\upsilon-1)(K//\Upsilon)+1,\dots,\upsilon(K//\Upsilon)$}
        \State \text{Call alg. \ref{alg:capped_vjp} for {\scriptsize$\Xi=\left( \vjp_{\mathbf{C}^t_k}(\frac{\diff l(\mathbf{o}^t)}{\diff \mathbf{y}_K^t}\otimes \mathbf{h}^t_k),\, \sum_{i=1}^{t}\vjp_{\mathbf{A}^i_k}(\frac{\diff l(\mathbf{o}^t)}{\diff \mathbf{y}_K^t}\blambda^{t,i}_k\otimes \mathbf{h}^{i-1}_k), \,\sum_{i=1}^{t}\vjp_{\mathbf{B}^i_k}(\frac{\diff l(\mathbf{o}^t)}{\diff \mathbf{y}_K^t}\blambda^{t,i}_k\otimes \hat{\mathbf{y}}^i_{k-1})\right )$}}
        \State \text{Compute: } $\frac{\diff L}{\diff \boldsymbol{\theta}}+= \Xi$
    \EndFor
    \For{Time step index $t=\Bar{T}+1,\dots,T$, layer index $k=(\upsilon-1)(K//\Upsilon)+1,\dots,\upsilon(K//\Upsilon)$}
        \State \begin{varwidth}[t]{\linewidth}
      \par\hskip\algorithmicindent Call alg. \ref{alg:capped_vjp} for $\Xi=\Biggl( \vjp_{\mathbf{C}^t_k}(\frac{\diff l(\mathbf{o}^t)}{\diff \mathbf{y}_K^t}\otimes \mathbf{h}^t_k),\, \sum_{i=t+1-\Bar{T}}^{t}\vjp_{\mathbf{A}^i_k}(\frac{\diff l(\mathbf{o}^t)}{\diff \mathbf{y}_K^t}\blambda^{t,i}_k\otimes \mathbf{h}^{i-1}_k),$
      \par\hskip\algorithmicindent $\sum_{i=t+1-\Bar{T}}^{t}\vjp_{\mathbf{B}^i_k}(\frac{\diff l(\mathbf{o}^t)}{\diff \mathbf{y}_K^t}\blambda^{t,i}_k\otimes \hat{\mathbf{y}}^i_{k-1}) \Biggl)$
      \end{varwidth}
        \State \text{Compute: } $\frac{\diff L}{\diff \boldsymbol{\theta}}+= \Xi$
        %\State \text{Release } $\Xi$
    \EndFor
    \State \text{\bf Return: $\frac{\diff L}{\diff \boldsymbol{\theta}}$}
\end{algorithmic}
\end{algorithm}

As shown in algorithm \ref{alg:capped_vjp}, to compute the $\vjp$s' for token index $t$ and ResNet index $k$, we only need $t,k,\diff l(\mathbf{o}^t)/\diff \mathbf{y}_K^t, \{\mathbf{h}_k^i\}_{i=0}^t,\mathbf{C}_k^t, \{\hat{\mathbf{y}}_{k-1}^i\}_{i=1}^t,\{\mathbf{A}_k^i\}_{i=2}^t$. 
To compute all the gradients for layer $k$, we only need $\mathbf{A}$, $\mathbf{h}$, and $\mathbf{C}$ from the $k$-th layer, and $\hat{\mathbf{y}}$ from the $k-1$-th layer. 
Therefore, we can divide the $K$ layers into $\Upsilon$ pieces, as shown in the appendix \ref{sec:dist}.

As the computations are fully independent and we compute the gradients using only data on local devices, we additionally distribute the model and the gradients, as shown in \autoref{tab:dist5}, where $\boldsymbol{\theta}_k$ represents the parameters of $\A_k$, $\B_k$, and $\C_k$, and $\mathrm{Gradient}_k$ represents the optimizer states for $\boldsymbol{\theta}_k$.

The complete training streamline is then as shown in algorithm \ref{alg:dist_capped_grad}. We fully distribute the activations, computations, gradients, and optimization states across $\Upsilon$ devices. While the forward evaluation pass results across different devices, as shown in algorithm \ref{alg:forward}, the computation of gradients is parallel across the $\Upsilon$ devices. This will speed up the training as the gradient computation takes most of the computation budget. We will also get a memory per GPU close to $\mathrm{Mem}/\Upsilon$, with $\mathrm{Mem}$ being the memory cost if we only have a single GPU. If we have $\Upsilon > K$ devices, we can further speed up the forward evaluation by first evaluating $\A$, $\B$, $\C$ in parallel, and then sequentially add them together on the distributed devices.
 
\subsection{Parallel computing}
\label{sec:parallel}
Adjoint sharding converts the sequential process of backpropagation gradient computation into individual independent $\vjp$s, allowing for parallel computation. We analyze the time and memory cost of $\vjp_{\boldsymbol{\A}^i_k}((\diff l^t/\diff \mathbf{y}_K^t)\blambda^{t,i}_k\otimes \mathbf{h}^{i-1}_k)$, $\vjp_{\boldsymbol{\B}^i_k}((\diff l^t/\diff \mathbf{y}_K^t)\blambda^{t,i}_k\otimes \hat{\mathbf{y}}^i_{k-1})$, and $\vjp_{\boldsymbol{\C}^t_k}((\diff l^t/\diff \mathbf{y}_K^t)\otimes \mathbf{h}^t_k)$. 

$\vjp$ has a similar time complexity as a forward pass, and a memory complexity of $\mathrm{bs}(|\boldsymbol{\theta}| + \mathbb{O})+|\boldsymbol{\theta}|$, where $\mathrm{bs}$ is the batch size, $\mathbb{O}$ is the number of elements in the network output, and $|\boldsymbol{\theta}|$ is the number of parameters \citep{novak2022fastfinitewidthneural}. We provide the memory and FLOPs required to compute the $\vjp$s in \autoref{tab:vjpMemFlop} \citep{nvidiamvp}. 

% Please add the following required packages to your document preamble:
% \usepackage{multirow}
\begin{table}[hbt!]
\scalebox{0.85}{
\begin{tabular}{lllll}
\toprule
\multicolumn{2}{l}{}                      & $\vjp_{\boldsymbol{\A}}$                                                                                                                                                                                       & $\vjp_{\boldsymbol{\B}}$                                                & $\vjp_{\boldsymbol{\C}}$                                                \\ \midrule
\multirow{2}{*}{Unstructured SSM} & Memory & $\mathrm{bs}(N^2+|\boldsymbol{\theta}_{\boldsymbol{\A}}|^*)+|\boldsymbol{\theta}_{\boldsymbol{\A}}|$ & $\mathrm{bs}(NP+|\boldsymbol{\theta}_{\boldsymbol{\B}}|^*)+|\boldsymbol{\theta}_{\boldsymbol{\B}}|$ & $\mathrm{bs}(NP+|\boldsymbol{\theta}_{\boldsymbol{\C}}|^*)+|\boldsymbol{\theta}_{\boldsymbol{\C}}|$ \\
                                  & FLOPs  & $\mathrm{bs}(N^2(2P+1))$                                                                                                                                                                                              & $\mathrm{bs}(NP(2P+1))$                                                        & $\mathrm{bs}(NP\times(2P+1))$                                                        \\
\multirow{2}{*}{Diagonal SSM}     & Memory & $\mathrm{bs}(N+|\boldsymbol{\theta}_{\boldsymbol{\A}}|^*)+|\boldsymbol{\theta}_{\boldsymbol{\A}}|$                                                      & $\mathrm{bs}(N+|\boldsymbol{\theta}_{\boldsymbol{\B}}|^*)+|\boldsymbol{\theta}_{\boldsymbol{\B}}|$  & $\mathrm{bs}(N+|\boldsymbol{\theta}_{\boldsymbol{\C}}|^*)+|\boldsymbol{\theta}_{\boldsymbol{\C}}|$ \\
                                  & FLOPs  & $\mathrm{bs}(N(2P+1))$               & $\mathrm{bs}(N(2P+1))$                                                         & $\mathrm{bs}(N(2P+1))$                                                        \\
\multirow{2}{*}{Scalar SSM}       & Memory & $\mathrm{bs}(1+|\boldsymbol{\theta}_{\boldsymbol{\A}}|^*)+|\boldsymbol{\theta}_{\boldsymbol{\A}}|$                                                                                                                                         & $\mathrm{bs}(N+|\boldsymbol{\theta}_{\boldsymbol{\B}}|^*)+|\boldsymbol{\theta}_{\boldsymbol{\B}}|$  & $\mathrm{bs}(N+|\boldsymbol{\theta}_{\boldsymbol{\C}}|^*)+|\boldsymbol{\theta}_{\boldsymbol{\C}}|$ \\
                                  & FLOPs  & $\mathrm{bs}(2P+1)$                                                                                                                                                                                                 & $\mathrm{bs}((N(2P+1))$                                                         & $\mathrm{bs}(N(2P+1))$   \\ \bottomrule                                                    
\end{tabular}
}
\caption{Memory and FLOPs required to compute the $\vjp$s. $|\boldsymbol{\theta}_{\boldsymbol{\A}}|^*$, $|\boldsymbol{\theta}_{\boldsymbol{\B}}|^*$, and $|\boldsymbol{\theta}_{\boldsymbol{\C}}|^*$ represents the number of elements of the biggest parameter vector of $\boldsymbol{\A}$, $\boldsymbol{\B}$, and $\boldsymbol{\C}$.}
\label{tab:vjpMemFlop}
\end{table}

We analyze training with a dataset containing contexts of lengths $T$, with $\Upsilon$ NVIDIA H100 GPUs, and performing computations in FP16. We use a selective diagonal SSM with $K$ layers, and each $\boldsymbol{\A}_k$, $\boldsymbol{\B}_k$, and $\boldsymbol{\C}_k$ network is a single-layer multi-layer perceptron (MLP). 

For each data point $\{\mathbf{x}^t\}_{t=1}^T$, we store $\{\mathbf{A}^t_k, \mathbf{C}^t_k, \mathbf{h}^t_k, \mathbf{y}^t_k\}_{(t,k)=(1,1)}^{(T,K)}$ and $\{\diff l(\mathbf{o}^t)/\diff \mathbf{y}_K^t\}_{t=1}^T$, which is $TK(2N+P)+TP$ FP16 numbers. We also save $\boldsymbol{\theta}_{\boldsymbol{\A}}$, $\boldsymbol{\theta}_{\boldsymbol{\B}}$, and $\boldsymbol{\theta}_{\boldsymbol{\C}}$, each taking $PN+N$ FP16 numbers. We need to store $T(2NK+PK+P)+3N(P+1)$ FP16 numbers before computing the  $\vjp$.

As computing all adjoint state sequences takes up to $N(2P+1)(1+T)T/2$ FLOPs, it takes $NP(1+T)/T$ FLOPs on average for each adjoint state. For $T$ large enough, $(1+T)/T\approx 1$, and we approximate the average FLOPs for each adjoint state with $NP$. Each $\vjp$ then takes $\mathrm{bs}(7NP+3N)$ FLOPs of computation. 

When computing with a selective diagonal SSM with $P=128$, $N=225$, and $\mathrm{bs}=8$, while storing and performing computations in FP16, computing $\vjp_{\boldsymbol{\A}}$, $\vjp_{\boldsymbol{\B}}$, and $\vjp_{\boldsymbol{\C}}$ each takes around $0.6\mathrm{MB}$ memory and $1798144$ FLOPs. The capacity of a modern GPU is mostly characterized by FLOPs/sec, which measures the computation speed; GPU memory bandwidth, which is the rate at which a GPU can move data between its memory and processing cores; GPU Memory, which is the amount of data a GPU can hold; and number of Multi-Instance GPU (MIG) instances, which is the number of fully isolated GPU instances with its own high-bandwidth memory, cache, and compute cores a GPU can host. 

An NVIDIA H100 Tensor Core GPU has a GPU memory bandwidth $3.35\mathrm{TB/s}$ and performs $1,979$ tera FP16 FLOPS per second. Therefore, the memory bandwidth allows computing $(3.35\mathrm{TB/s})/0.6\mathrm{MB}=5.58\times 10\mathrm{E}6$ batches of $\vjp$s per second, and the computing speed allows computing $(1979\mathrm{tera/s})/1798144=3.76\times 1.1\mathrm{E}9$ batches of $\vjp$s per second. At the same time, since the H100 GPU has $80\mathrm{GB}$ memory, it can hold up to $80\mathrm{GB}/(0.6\mathrm{MB}/\vjp)=133$ batches of $\vjp$s at the same time if we do not consider any memory overhead. 
As each H100 GPU can hold up to $7$ instances in parallel, we perform the adjoint sharding algorithm with $7\Upsilon$ instances, offering as much as a 56x speedup on one AWS P4 instance (8 H100 GPUs). Such speedup cannot be achieved for backpropagation because of its sequential nature.
%We assume an SSM model with $K=300$ layers, leading to a model with $30\mathrm{M}$ parameters. 

%\xx{analyze the memory cost of each block of the $\vjp$ and compare it to backpropagation}

\paragraph{Limitation}
The adjoint sharding method provides an alternative method of computing gradients to backpropagation. While we analytically proved that the gradients computed from adjoint sharding equals to that from backpropagation, adjoint sharding suffer from a time complexity polynomial regarding the training context length when computing equivalent gradients. We provided the truncated adjoint sharding as a linear time complexity alternative, and leave the analysis of its convergence and further improvements on it for future works. We also provided a distributed and parallel computing algorithm for performing adjoint sharding. However, the overhead of naïve implementation of such algorithm with multi-threading or multiprocessing overweights the speedups when the training context length is small. We leave efficient implementation of the parallel algorithm on a CUDA kernel for future work.

\paragraph{Conclusion}
We introduced adjoint sharding, a distributed and parallel computing algorithm, to facilitate training of LLMs on long contexts. Unlike the sequential backpropagation, the adjoint sharding computes gradients of each LLM layer against each token independently through vector-Jacobian product, allowing for parallel computation. To avoid the limitation of $\vjp$s increasing polynomially regarding context length, we propose truncated adjoint sharding to focus on important gradients. We analyzed the memory and FLOP cost of each computation block in adjoint sharding and proposed a method to accelerate it through parallel computing. Empirical results suggest orders of magnitude of memory reduction in training while maintaining the same training results as backpropagation. 

\clearpage
\bibliography{references}
\bibliographystyle{plainnat}  

\appendix
\section{Appendix}

\subsection{Proof for proposition \ref{prop:onessm}}
\label{sec:proofOnessm}
\begin{proof}
\label{proof:onessm}
    Define $\partial \Tilde{\mathbf{y}}/\partial \mathbf{h}^t=\Tilde{\mathbf{y}}^t_{\mathbf{h}^t}$, $\partial \Tilde{\mathbf{h}}^t/\partial \mathbf{h}^{t-1}=\Tilde{\mathbf{h}}^t_{\mathbf{h}^{t-1}}$, and $\partial \Tilde{\mathbf{y}}/\partial \boldsymbol{\theta}=\Tilde{\mathbf{y}}^t_{\boldsymbol{\theta}}$, $\partial \Tilde{\mathbf{h}}^t/\partial \boldsymbol{\theta}=\Tilde{\mathbf{h}}^t_{\boldsymbol{\theta}}$, by plugging in the expression for $\Tilde{\mathbf{y}}^t$ from \autoref{sec:resnet}, proposition \ref{prop:adjoint} states that 
\begin{equation*}
    \frac{\diff \Tilde{\mathbf{y}}^t}{\diff \boldsymbol{\theta}} = \Tilde{\mathbf{y}}^t_{\mathbf{h}^t} \left[ (\prod_{i=1}^{t-1}\mathbf{h}^{t-i+1}_{\mathbf{h}^{t-i}})\mathbf{h}^1_{\boldsymbol{\theta}}+(\prod_{i=1}^{t-2}\mathbf{h}^{t-i+1}_{\mathbf{h}^{t-i}})\mathbf{h}^2_{\boldsymbol{\theta}}+\dots+\mathbf{h}^t_{\mathbf{h}^{t-1}}\mathbf{h}^{t-1}_{\boldsymbol{\theta}}+\mathbf{h}^t_{\boldsymbol{\theta}} \right] +\Tilde{\mathbf{y}}^t_{\boldsymbol{\theta}}.
\end{equation*}
In the context of $\ssm$, we have:
\begin{equation}
    \mathbf{h}^t=\mathbf{A}^t\mathbf{h}^{t-1}+\mathbf{B}^t\hat{\mathbf{x}}^t, \mathbf{h}^t_{\mathbf{h}^{t-1}} = \mathbf{A}^t, \mathbf{h}^t_{\boldsymbol{\theta}}=\mathbf{A}^t_{\boldsymbol{\theta}}\mathbf{h}^{t-1}+\mathbf{B}^t_{\boldsymbol{\theta}}\hat{\mathbf{x}}^t, \tilde{\mathbf{y}}^t=\mathbf{C}^t\mathbf{h}^t, \tilde{\mathbf{y}}^t_{\mathbf{h}^t}=\mathbf{C}^t, \tilde{\mathbf{y}}^t_{\boldsymbol{\theta}}=\mathbf{C}^t_{\boldsymbol{\theta}}\mathbf{h}^t.
\end{equation}
Plugging in these relations, we get:
\begin{equation}
    \frac{\diff \tilde{\mathbf{y}}^t}{\diff\boldsymbol{\theta}} = \mathbf{C}^t \left[ (\prod_{i=1}^{t-1}\mathbf{A}^{t+1-i})\mathbf{h}^1_{\boldsymbol{\theta}}+(\prod_{i=1}^{t-2}\mathbf{A}^{t+1-i})\mathbf{h}^2_{\boldsymbol{\theta}}+\dots+(\prod_{i=1}^2\mathbf{A}^{t+1-i})\mathbf{h}^{t-2}_{\boldsymbol{\theta}}+\mathbf{A}^t\mathbf{h}^{t-1}_{\boldsymbol{\theta}}+\mathbf{h}^t_{\boldsymbol{\theta}} \right] +\tilde{\mathbf{y}}^t_{\boldsymbol{\theta}}.
\end{equation}

Define the adjoint state $\blambda^{t,\tau}=\mathbf{C}^t(\prod_{i=1}^{t-\tau}\mathbf{A}^{t+1-i})$, we have
\begin{align*}
    \frac{\diff \tilde{\mathbf{y}}^t}{\diff\boldsymbol{\theta}} &= \blambda^{t,1}\mathbf{h}^1_{\boldsymbol{\theta}}+\blambda^{t,2}\mathbf{h}^2_{\boldsymbol{\theta}}+\dots+\blambda^{t,t-1}
\mathbf{h}^{t-1}_{\boldsymbol{\theta}}+\blambda^{t,t}\mathbf{h}^t_{\boldsymbol{\theta}}+\tilde{\mathbf{y}}^t_{\boldsymbol{\theta}}
\end{align*}
Therefore, we have
\begin{align*}
    \frac{\diff l^t}{\diff\boldsymbol{\theta}} &= \frac{\diff l^t}{\diff \mathbf{y}^t}\frac{\diff (\tilde{\mathbf{y}}^t+\hat{\mathbf{x}}^t)}{\diff\boldsymbol{\theta}}\\%\textrm{\ka{hmmm, i don't understand this piece}}\\
    &= \frac{\diff l^t}{\diff \mathbf{y}^t}\frac{\diff \tilde{\mathbf{y}}^t}{\diff\boldsymbol{\theta}}\\
    &= \frac{\diff l^t}{\diff \mathbf{y}^t} [\blambda^{t,1}\mathbf{h}^1_{\boldsymbol{\theta}}+\blambda^{t,2}\mathbf{h}^2_{\boldsymbol{\theta}}+\dots+\blambda^{t,t-1}\mathbf{h}^{t-1}_{\boldsymbol{\theta}}+\blambda^{t,t}\mathbf{h}^t_{\boldsymbol{\theta}}+\tilde{\mathbf{y}}^t_{\boldsymbol{\theta}}]
\end{align*}
%\ka{okay i think i understand. Since here there is only one layer, the output is the sum of the input + the output of SSM which is $\tilde y^t$ yes? in that case, i'd work with $x$ rather than $\hat x$ since with one layer, the input is really just the original input, but discard the comment if it does not make sense}
%% reply to ka: we should use $\hat x$ rather than $x$ as we apply a normalization before putting it through the networks

Plug in everything, we have
\begin{align*}
    \frac{\diff l^t}{\diff\boldsymbol{\theta}} &= \frac{\diff l^t}{\diff \mathbf{y}^t}[\blambda^{t,1}(\mathbf{A}^1_{\boldsymbol{\theta}}\mathbf{h}^0+\mathbf{B}^1_{\boldsymbol{\theta}}\hat{\mathbf{x}}^1)+\blambda^{t,2}(\mathbf{A}^2_{\boldsymbol{\theta}}\mathbf{h}^1+\mathbf{B}^2_{\boldsymbol{\theta}}\hat{\mathbf{x}}^2)+\dots+\blambda^{t,t}(\mathbf{A}^t_{\boldsymbol{\theta}}\mathbf{h}^{t-1}+\mathbf{B}^t_{\boldsymbol{\theta}}\hat{\mathbf{x}}^t)+\mathbf{C}^t_{\boldsymbol{\theta}}\mathbf{h}^t\\
    &= \left[ \sum_{i=1}^{t}\frac{\diff l^t}{\diff \mathbf{y}^t}\blambda^{t,i}(\mathbf{A}^i_{\boldsymbol{\theta}}\mathbf{h}^{i-1}+\mathbf{B}^i_{\boldsymbol{\theta}}\hat{\mathbf{x}}^i) \right] +\frac{\diff l^t}{\diff \mathbf{y}^t}\mathbf{C}^t_{\boldsymbol{\theta}}\mathbf{h}^t\\
    &= \left[ \sum_{i=1}^{t}\vjp_{\boldsymbol{\A}^i}(\frac{\diff l^t}{\diff \mathbf{y}^t}\blambda^{t,i}\otimes \mathbf{h}^{i-1})+\vjp_{\boldsymbol{\B}^i}(\frac{\diff l^t}{\diff \mathbf{y}^t}\blambda^{t,i}\otimes \hat{\mathbf{x}}^i) \right] +\vjp_{\boldsymbol{\C}^t}(\frac{\diff l^t}{\diff \mathbf{y}^t}\otimes \mathbf{h}^t)
\end{align*}
where we define $\vjp_{NN^i}(v)=v\cdot NN_{\boldsymbol{\theta}}(\mathrm{Input}^i)$, with $\boldsymbol{\theta}$ being $NN$'s parameters and $i$ being the index of $\mathrm{Input}$. Now, as $\vjp_{\boldsymbol{\A}^i}(\frac{\diff l^t}{\diff \mathbf{y}^t}\blambda^{t,i}\otimes \mathbf{h}^{i-1})$, $\vjp_{\boldsymbol{\B}^i}(\frac{\diff l^t}{\diff \mathbf{y}^t}\blambda^{t,i}\otimes \hat{\mathbf{x}}^i)$, and $\vjp_{\boldsymbol{\C}^t}(\frac{\diff l^t}{\diff \mathbf{y}^t}\otimes \mathbf{h}^t)$are separate, we have 
\begin{equation}
    \frac{\diff l^t}{\diff\boldsymbol{\theta}} = \left[ \sum_{i=1}^{t}\vjp_{\boldsymbol{\A}^i}(\frac{\diff l^t}{\diff \mathbf{y}^t}\blambda^{t,i}\otimes \mathbf{h}^{i-1})\right ] \oplus \left [ \sum_{i=1}^{t}\vjp_{\boldsymbol{\B}^i}(\frac{\diff l^t}{\diff \mathbf{y}^t}\blambda^{t,i}\otimes \hat{\mathbf{x}}^i) \right ] \oplus \vjp_{\boldsymbol{\C}^t}(\frac{\diff l^t}{\diff \mathbf{y}^t}\otimes \mathbf{h}^t),
\end{equation}
where $\oplus$ is vector concatenation.
\end{proof}

\subsection{Proof for proposition \ref{prop:multssm}}
\label{sec:proofMultssm}
\begin{proof}
\label{proof:multssm}
First, using the structure of ResNet, we have
\begin{align*}
    \frac{\diff L}{\diff \boldsymbol{\theta}} &= \sum_{t=1}^T\frac{\diff l^t}{\diff \mathbf{y}_K^t}\frac{\diff \mathbf{y}_K^t}{\diff \boldsymbol{\theta}}\\
    &= \sum_{t=1}^T\frac{\diff l^t}{\diff \mathbf{y}_K^t}\frac{\diff (\mathbf{y}_0^t+\sum_{k=1}^K \tilde{\mathbf{y}}_k^t)}{\diff \boldsymbol{\theta}}\\
    &= \sum_{t=1}^T\frac{\diff l^t}{\diff \mathbf{y}_K^t}\sum_{k=1}^K\frac{\diff \tilde{\mathbf{y}}_k^t}{\diff \boldsymbol{\theta}}\\
    &= \sum_{t=1}^T\sum_{k=1}^K\frac{\diff l^t}{\diff \mathbf{y}_K^t}\frac{\diff \tilde{\mathbf{y}}_k^t}{\diff \boldsymbol{\theta}}
\end{align*}
from proposiiton \ref{prop:onessm}, we have proven that for a single SSM model, we have
\begin{equation*}
    \frac{\diff l^t}{\diff\boldsymbol{\theta}} = \left[ \sum_{i=1}^{t}\vjp_{\boldsymbol{\A}^i}(\frac{\diff l^t}{\diff \mathbf{y}^t}\blambda^{t,i}\otimes \mathbf{h}^{i-1})\right ] \oplus \left [ \sum_{i=1}^{t}\vjp_{\boldsymbol{\B}^i}(\frac{\diff l^t}{\diff \mathbf{y}^t}\blambda^{t,i}\otimes \hat{\mathbf{x}}^i) \right ] \oplus \vjp_{\boldsymbol{\C}^t}(\frac{\diff l^t}{\diff \mathbf{y}^t}\otimes \mathbf{h}^t),
\end{equation*}
so for the ResNet model, we have
\begin{equation}
\label{eqn:resnet}
\begin{split}
    \frac{\diff L}{\diff \boldsymbol{\theta}} &= \sum_{t=1}^T\sum_{k=1}^K\frac{\diff l^t}{\diff \mathbf{y}_K^t}\frac{\diff \tilde{\mathbf{y}}_k^t}{\diff \boldsymbol{\theta}}\\
    &=\sum_{t=1}^T\sum_{k=1}^K \left\{ \left[ \sum_{i=1}^{t}\vjp_{\boldsymbol{\A}^i_k}(\frac{\diff l^t}{\diff \mathbf{y}_K^t}\blambda^{t,i}_k\otimes \mathbf{h}^{i-1}_k)\right] \oplus \left[\sum_{i=1}^{t}\vjp_{\boldsymbol{\B}^i_k}(\frac{\diff l^t}{\diff \mathbf{y}_K^t}\blambda^{t,i}_k\otimes \hat{\mathbf{x}}^i_k) \right] \oplus \vjp_{\boldsymbol{\C}^t_k}(\frac{\diff l^t}{\diff \mathbf{y}_K^t}\otimes \mathbf{h}^t_k) \right\}\\
    &= \left(\sum_{t=1}^T\sum_{k=1}^K \vjp_{\boldsymbol{\C}^t_k}(\frac{\diff l^t}{\diff \mathbf{y}_K^t}\otimes \mathbf{h}^t_k)\right)\\
    &\oplus \left(\sum_{t=1}^T\sum_{k=1}^K \sum_{i=1}^{t}\vjp_{\boldsymbol{\A}^i_k}(\frac{\diff l^t}{\diff \mathbf{y}_K^t}\blambda^{t,i}_k\otimes \mathbf{h}^{i-1}_k)\right)\\
    &\oplus \left(\sum_{t=1}^T\sum_{k=1}^K \sum_{i=1}^{t}\vjp_{\boldsymbol{\B}^i_k}(\frac{\diff l^t}{\diff \mathbf{y}_K^t}\blambda^{t,i}_k\otimes \hat{\mathbf{x}}^i_k)\right)\\
    &= \left(\sum_{t=1}^T\sum_{k=1}^K \vjp_{\boldsymbol{\C}^t_k}(\frac{\diff l^t}{\diff \mathbf{y}_K^t}\otimes \mathbf{h}^t_k)\right)\\ 
    &\oplus \left(\sum_{t=1}^T\sum_{k=1}^K \sum_{i=1}^{t}\vjp_{\boldsymbol{\A}^i_k}(\frac{\diff l^t}{\diff \mathbf{y}_K^t}\blambda^{t,i}_k\otimes \mathbf{h}^{i-1}_k)\right)\\
    &\oplus \left(\sum_{t=1}^T\sum_{k=1}^K \sum_{i=1}^{t}\vjp_{\boldsymbol{\B}^i_k}(\frac{\diff l^t}{\diff \mathbf{y}_K^t}\blambda^{t,i}_k\otimes \hat{\mathbf{y}}^i_{k-1})\right)
\end{split}
\end{equation}
where the input to $\vjp_{\boldsymbol{\C}^t_k}(\frac{\diff l^t}{\diff \mathbf{y}_K^t}\otimes \mathbf{h}^t_k),\, \vjp_{\boldsymbol{\A}^i_k}(\frac{\diff l^t}{\diff \mathbf{y}_K^t}\blambda^{t,i}_k\otimes \mathbf{h}^{i-1}_k), \,\mathrm{and}\,\vjp_{\boldsymbol{\B}^i_k}(\frac{\diff l^t}{\diff \mathbf{y}_K^t}\blambda^{t,i}_k\otimes \hat{\mathbf{y}}^i_{k-1})$ are computed with the k-th $\ssm$ and the $\hat{\mathbf{x}}_k^i=\hat{\mathbf{y}}_{k-1}^i=\mathrm{RMSNorm}(\mathbf{y}_{k-2}^i+\ssm_{k-1}(\hat{\mathbf{Y}}_{k-2})^i)$ (the normalized output sequence of the (k-1)-th $\ssm$), and the adjoint state $\blambda^{t,\tau}_k=\mathbf{C}^t_k(\prod_{i=1}^{t-\tau}\mathbf{A}_k^{t+1-i})$.
    
\end{proof}

\subsection{Proof of concept for VJP}
As a proof of concept of why $(\diff l^t/\diff \mathbf{y}^t) \mathbf{C}^t_{\boldsymbol{\theta}} \mathbf{h}^t$ can computed with $\vjp$, we present an explicit and simple example. 
We have $\mathbf{y}=[y_1,y_2]$, $\mathbf{h}=[h_1,h_2,h_3]$, $\boldsymbol{\theta}=\Vec{\boldsymbol{\theta}}$. 
We then have 
\[
\frac{dl}{d\mathbf{y}}=
\begin{bmatrix}
    l_{y_1} & l_{y_2}
\end{bmatrix}\in\R^{1\times P}
\]
\[
\mathbf{C}_{\boldsymbol{\theta}}=
\begin{bmatrix}
    C_{11}^{\Vec{\boldsymbol{\theta}}} & C_{12}^{\Vec{\boldsymbol{\theta}}} & C_{13}^{\Vec{\boldsymbol{\theta}}}\\
    C_{21}^{\Vec{\boldsymbol{\theta}}} & C_{22}^{\Vec{\boldsymbol{\theta}}} & C_{23}^{\Vec{\boldsymbol{\theta}}}\\
\end{bmatrix}\in\R^{P\times N\times |\boldsymbol{\theta}|}
\]
\[
\mathbf{h} = 
\begin{bmatrix}
    h_1\\
    h_2\\
    h_3
\end{bmatrix} \in\R^{N\times 1}
\]
With each $C_{ij}^{\Vec{\boldsymbol{\theta}}}=[\partial C_{ij}/\partial \boldsymbol{\theta}_1, \dots,\partial C_{ij}/\partial \boldsymbol{\theta}_{|\boldsymbol{\theta}|}]\in\R^{|\boldsymbol{\theta}|}$.
We have 

\begin{align*}
    \frac{\diff l}{\diff y}\mathbf{C}_{\boldsymbol{\theta}}\mathbf{h} 
    &= C_{11}^{\Vec{\boldsymbol{\theta}}}l_{y_1}h_1+C_{21}^{\Vec{\boldsymbol{\theta}}}l_{y_2}h_1+C_{12}^{\Vec{\boldsymbol{\theta}}}l_{y_1}h_2+C_{22}^{\Vec{\boldsymbol{\theta}}}l_{y_2}h_2+C_{13}^{\Vec{\boldsymbol{\theta}}}l_{y_1}h_3+C_{23}^{\Vec{\boldsymbol{\theta}}}l_{y_2}h_3\\
    &= [
        l_{y_1}h_1 \; l_{y_1}h_2 \; l_{y_1}h_3 \; l_{y_2}h_1 \; l_{y_2}h_2 \; l_{y_2}h_3
        ]
        \cdot 
        [
        C_{11}^{\Vec{\boldsymbol{\theta}}} \; C_{12}^{\Vec{\boldsymbol{\theta}}} \; C_{13}^{\Vec{\boldsymbol{\theta}}}
        C_{21}^{\Vec{\boldsymbol{\theta}}} \; C_{22}^{\Vec{\boldsymbol{\theta}}} \; C_{23}^{\Vec{\boldsymbol{\theta}}}
        ]\\
    &= \mathrm{sum}\left( (\begin{bmatrix}
        l_{y_1}\\
        l_{y_2}
        \end{bmatrix}
        \otimes 
        \begin{bmatrix}
        h_1 & h_2 & h_3
        \end{bmatrix})
        \circ
        \begin{bmatrix}
        C_{11}^{\Vec{\boldsymbol{\theta}}} & C_{12}^{\Vec{\boldsymbol{\theta}}} & C_{13}^{\Vec{\boldsymbol{\theta}}}\\
        C_{21}^{\Vec{\boldsymbol{\theta}}} & C_{22}^{\Vec{\boldsymbol{\theta}}} & C_{23}^{\Vec{\boldsymbol{\theta}}}\\
        \end{bmatrix}
        \right)
\end{align*}
where $\cdot$ is vector dot product, $\otimes$ is vector outer product, $\circ$ is element-wise product, and $\mathrm{sum}$ means summing all elements in a matrix.

\subsection{Distributed tensors' locations}
\label{sec:dist}
We provide the specific location for each tensors in distributed training:
\begin{table}[hbt!]
\caption{Tensors stored on each GPU, part 1.}
\begin{tabular}{l|ll}
GPU index             & $\diff l(\mathbf{o}^t)/\diff y_K^t$ & $h_k^t$  \\ \midrule
$\upsilon=1$          & $t=1,\dots,T$              & $t=1,\dots,T;\, k=1,\dots K//\Upsilon$ \\
$\upsilon=2$          & $t=1,\dots,T$              & $t=1,\dots,T;\, k=K//\Upsilon+1,\dots,2(K//\Upsilon)$ \\
$\dots$               & $\dots$                    & $\dots$ \\
$\upsilon=\Upsilon-1$ & $t=1,\dots,T$              & $t=1,\dots,T;\, k=(\Upsilon-2)(K//\Upsilon)+1,\dots,(\Upsilon-1)(K//\Upsilon)$ \\
$\upsilon=\Upsilon$   & $t=1,\dots,T$              & $t=1,\dots,T;\, k=(\Upsilon-1)(K//\Upsilon)+1,\dots,K$ 
\end{tabular}
\label{tab:dist1}
\end{table}

\begin{table}[hbt!]
\caption{Tensors stored on each GPU, part 2.}
\begin{tabular}{l|l}
GPU index             & $C_k^t$ \\ \midrule
$\upsilon=1$          & $t=1,\dots,T;\, k=1,\dots K//\Upsilon$ \\
$\upsilon=2$ & $t=1,\dots,T;\, k=K//\Upsilon+1,\dots,2(K//\Upsilon)$ \\
$\dots$ & $\dots$ \\
$\upsilon=\Upsilon-1$ & $t=1,\dots,T$ \\
$\upsilon=\Upsilon$   & $t=1,\dots,T;\, k=(\Upsilon-1)(K//\Upsilon)+1,\dots,K$
\end{tabular}
\label{tab:dist2}
\end{table}

\begin{table}[hbt!]
\caption{Tensors stored on each GPU, part 3.}
\begin{tabular}{l|l}
GPU index & $\hat{y}_k^t$  \\ \midrule
$\upsilon=1$ & $t=1,\dots,T;\, k=0,\dots K//\Upsilon-1$ \\
$\upsilon=2$ & $t=1,\dots,T;\, k=K//\Upsilon,\dots,2(K//\Upsilon)-1$ \\
$\dots$ & $\dots$ \\
$\upsilon=\Upsilon-1$ & $t=1,\dots,T;\, k=(\Upsilon-2)(K//\Upsilon),\dots,(\Upsilon-1)(K//\Upsilon)-1$ \\
$\upsilon=\Upsilon$   & $t=1,\dots,T;\, k=(\Upsilon-1)(K//\Upsilon),\dots,K-1$
\end{tabular}
\label{tab:dist3}
\end{table}

\begin{table}[hbt!]
\caption{Tensors stored on each GPU, part 4.}
\begin{tabular}{l|l}
GPU index & $A_k^t$ \\ \midrule
$\upsilon=1$ & $t=2,\dots,T;\, k=1,\dots K//\Upsilon$ \\
$\upsilon=2$ & $t=2,\dots,T;\, k=K//\Upsilon+1,\dots,2(K//\Upsilon)$ \\
$\dots$ & $\dots$\\
$\upsilon=\Upsilon-1$ & $t=2,\dots,T;\, k=(\Upsilon-2)(K//\Upsilon)+1,\dots,(\Upsilon-1)(K//\Upsilon)$ \\
$\upsilon=\Upsilon$ & $t=2,\dots,T;\, k=(\Upsilon-1)(K//\Upsilon)+1,\dots,K$
\end{tabular}
\label{tab:dist4}
\end{table}

\begin{table}[hbt!]
\caption{Tensors stored on each GPU, part 5.}
\begin{tabular}{l|ll}
GPU index & $\boldsymbol{\theta}_k$ & $\mathrm{Gradient}_k$\\ \midrule
$\upsilon=1$ & $k=1,\dots K//\Upsilon$ & $k=1,\dots K//\Upsilon$ \\
$\upsilon=2$ & $k=K//\Upsilon+1,\dots,2(K//\Upsilon)$ & $k=K//\Upsilon+1,\dots,2(K//\Upsilon)$ \\
$\dots$ & $\dots$ & $\dots$\\
$\upsilon=\Upsilon-1$ & $k=(\Upsilon-2)(K//\Upsilon)+1,\dots,(\Upsilon-1)(K//\Upsilon)$ & $k=(\Upsilon-2)(K//\Upsilon)+1,\dots,(\Upsilon-1)(K//\Upsilon)$ \\
$\upsilon=\Upsilon$ & $k=(\Upsilon-1)(K//\Upsilon)+1,\dots,K$ & $k=(\Upsilon-1)(K//\Upsilon)+1,\dots,K$              
\end{tabular}
\label{tab:dist5}
\end{table}
\end{document}